\pdfoutput=1
\documentclass{article}
\usepackage{iclr2027_conference,times}

\usepackage{amsmath,amssymb,amsthm}

\usepackage{booktabs}
\usepackage{multirow}
\usepackage{array}
\usepackage{algorithm}
\usepackage{algorithmic}
\usepackage{graphicx}
\usepackage{float}
\usepackage[section]{placeins}   %

\usepackage{xcolor}
\usepackage{tikz}
\usetikzlibrary{arrows.meta,decorations.pathreplacing}
\definecolor{accent}{HTML}{15568C}
\definecolor{accentlt}{HTML}{4E8FC0}
\definecolor{c1}{HTML}{7FB3DA}
\definecolor{c2}{HTML}{E9AE58}
\definecolor{c3}{HTML}{86C67E}
\definecolor{c4}{HTML}{C58BC7}
\definecolor{c5}{HTML}{E88585}
\definecolor{c6}{HTML}{ADADAD}
\usepackage{xr-hyper}   %
\usepackage{hyperref}
\usepackage{url}

\newif\ifwithsupp
\withsupptrue

\ifwithsupp\else
\fi

\newcommand{\method}{RaFL}
\usepackage[table]{xcolor} %

\title{RaFL: An Empirical Study of Random Filter Losses for Image Reconstruction}

\author{Ruibo Wang$^{1}$, Fengbei Liu$^{2}$, Aleksei Tiulpin$^{3}$, Mert R. Sabuncu$^{2,3}$, Heejong Kim$^{3}$\\
$^{1}$TU Delft \quad $^{2}$Cornell Tech \quad $^{3}$Weill Cornell Medicine}

\iclrfinalcopy  %
\begin{document}
\maketitle
\lhead{Preprint. Under review.}  %

\begin{abstract}
Pixel losses train image reconstruction networks toward a point statistic of the target distribution, often averaging away fine textures under uncertainty. Perceptual losses alleviate this issue using pretrained features, but those features are inherently domain-specific. We study a random filter loss (\method{}) that compares prediction and target through filters with random weights, avoiding both pretrained features and task-specific loss engineering. \method{} combines random convolutions, random Fourier features, and random wavelet filters, automatically balancing their contributions through gradient-norm matching, eliminating manual loss-weight tuning and grid search. We systematically investigate three design choices: filter family, weight resampling schedule, and weight initialization.
Based on validation PSNR and stability across random seeds, we select the configuration that uses all three filter families, redraws filters every optimization step, and employs Gaussian initialization. We evaluate \method{} on three 2D and two 3D image reconstruction tasks. \method{} outperforms a pixel loss on three tasks and matches it on two, and demonstrates stable performance across random seeds on all five tasks. %
\method{} adds less than $2\%$ to the backbone's forward FLOPs and no inference cost, requiring $182\times$ fewer loss FLOPs than a VGG19 perceptual loss. These findings suggest that \method{} is a simple, tuning-free, and domain-agnostic loss objective for 2D and 3D image reconstruction.
\end{abstract}

\section{Introduction}
\label{sec:intro}

A pixel loss combined with a perceptual loss is a common setup for image reconstruction tasks, as a solution to the averaging behavior of pixel loss alone.
However, perceptual losses require a pretrained network, and the pretrained features are tied to the domain they were trained on. This limits their use on inputs from other domains, e.g., 3D medical images~\citep{pfaehler2025untrained}.
Fixed filters that require no learning, such as a gradient penalty \citep{mathieu2015deep}, a Laplacian edge term \citep{jiang2020multi}, a wavelet-domain discrepancy \citep{korkmaz2024training}, or a Fourier-domain one \citep{jiang2021focal}, have demonstrated effectiveness across various image reconstruction tasks.
However, their filter responses are fixed, and the selection of filters and weights are typically determined through ablations or expert knowledge, often requiring task-specific hyperparameter tuning.

Untrained network based losses that avoid the designing for filter choices have proven to be effective in image reconstruction tasks~\citep{lempitsky2018deep,heckel2019deep}, image generation~\citep{he2016powerful,ustyuzhaninov2017does}, and image segmentation~\citep{liu2021generic}.
\citet{zhou2023training} and \citet{pfaehler2025untrained} apply random weight networks as optimization functions, achieving strong results on 2D and 3D image reconstruction tasks, respectively.
However, how to design the random filter loss itself remains underexplored. \citet{liu2021generic} find that depth matters more than pretraining as a prior, while \citet{pfaehler2025untrained} find that a small network performs best as a loss function, and since each conclusion comes from experiments conducted in a specific setting, it is unclear which holds more broadly across both 2D and 3D.
We use the term random filter loss broadly to encompass objectives based on either explicit random filters or untrained random networks.

In this paper, we propose \method{} (pronounced "raffle"), a random filter loss built not from a convolutional or transformer architecture, but from simple, untrained filter families that complement each other.
\method{} uses a fixed loss formulation, balances its filter families automatically by gradient-norm matching, and preconditions each update with MAdam \citep{liu2026madam}.
Through empirical experiments, we answer three questions: which filters to use, how often to redraw their weights, and which initialization method to use.
We select a default configuration on SIDD for mean validation PSNR and cross-seed stability, combining all three filter families with per-step Gaussian redraws. With this configuration, we evaluate \method{} across five tasks: real-noise denoising (2D, SIDD), low-light enhancement (2D, LOL), pan-sharpening (2D, WorldView-2), super-resolution (3D, OASIS3), and MR angiography denoising (3D, IXI). \method{} adds no learnable parameters, is not needed at inference. It costs under two percent of the forward pass of the image reconstruction network it trains, and it needs $182\times$ fewer loss FLOPs than an untrained VGG19 perceptual loss. \method{} achieves PSNR and SSIM that are competitive with or better than those of the baselines across all five tasks (Tables~\ref{tab:apply2d} and \ref{tab:apply3d}). Its gains in perceptual metrics depend on the task and the backbone (Sec.~\ref{sec:apply}, App.~\ref{sec:transfer-ablations}).
\section{Related Work}
\label{sec:related}

\paragraph{Perceptual losses.} Pixel losses, such as $\ell_1$ and $\ell_2$, are the dominant choice of loss functions for various image reconstruction tasks.
Beyond the pixel losses, a perceptual loss compares prediction and target in the pretrained network features on natural images \citep{johnson2016perceptual}, LPIPS calibrates such features against human judgments of similarity \citep{zhang2018unreasonable}, and DISTS combines a structure and a texture comparison of them so that the distance tolerates resampled texture \citep{ding2020image}.
An adversarial objective trains its critic alongside the model.
These trained perceptual metrics carry the data they were trained on. Their features encode the inductive biases of natural images and are not calibrated for other modalities \citep{pfaehler2025untrained}. In 3D medical imaging, the pretrained encoders that exist are each tied to a specific modality and anatomy \citep{hatamizadeh2021swin}, so they do not transfer to new modalities or anatomies.
Structural losses such as SSIM and its multi-scale form \citep{wang2004image,wang2003multiscale} are often treated as a perceptual alternative in restoration \citep{zhao2016loss}, however, they compare only fixed local statistics inside a window at one or a few scales, making them a pixel loss with a larger window rather than a feature loss.

\paragraph{Handcrafted filter losses.}
Handcrafted filter terms have been proposed as a training-free alternative.
Image gradient based filters have been used for video prediction~\citep{mathieu2015deep}, image synthesis~\citep{yu2019ea} and image denoising~\citep{liu2020gradnet}.
Another line of work includes loss functions based on the Laplacian~\citep{jiang2020multi}, spectral-spatial structure~\citep{choi2020s3}, wavelet-domain~\citep{korkmaz2024training}, and frequency-domain~\citep{jiang2021focal,li2022transformer}.
Since no training is required, the computational cost of these losses is relatively negligible.
However, each response is fixed and applied identically regardless of the data.
Also, the selection and combination of filters are made a priori and are task-specific, requiring domain expertise to determine which filters to penalize and weight more heavily.

\paragraph{Untrained perceptual/filter losses.}
The effectiveness of untrained convolutional neural network (CNN) as a prior has been shown in various image reconstruction tasks~\citep{lempitsky2018deep,heckel2019deep}.
A randomly weighted CNN also have shown the utility in achieving image representations for generative tasks~\citep{he2016powerful} and synthesizing realistic textures~\citep{ustyuzhaninov2017does}.
\citet{liu2021generic} showed that an untrained CNN used as a perceptual loss performs comparably to a pretrained one on semantic segmentation, depth estimation, and instance segmentation, and reported that network structure matters more than training.
\citet{zhou2023training} proposed four random-weight filter prototypes as an image restoration loss constraint a central-difference convolution (CDC) \citep{yu2020searchingcentraldifferenceconvolutional}, zero-order reverse (reverse) filtering \citep{tao2017zero}, an invertible network (INN) \citep{dinh2014nice} and a Taylor expansion (Taylor). Each is added to a pixel loss and the weight redrawn per epoch, and showed that all four consistently improve performance over baseline.
The most closely related work is an untrained perceptual loss (uPL) \citet{pfaehler2025untrained}, that uses a small CNN with no pixel term as the loss. They apply the loss to 3D MR denoising of brain vessels and plant roots, varying the loss network's depth, kernel size, pooling and initialization. They report that kernel size matters little, pooling helps on one dataset but hurts on the other, and shallow networks win. Three to five layers match or beat an untrained VGG19, contrary to \citet{liu2021generic}, who found depth mattered.

These related works demonstrate the strong performance of random filter losses, but they remain narrower in scope. \citet{pfaehler2025untrained} use untrained networks without reweighting as a perceptual loss for 3D MR denoising of line-like structures, and \citet{zhou2023training} use random-weight networks built from a hand-chosen prototypes for image restoration with a pixel loss. In contrast, \method{} works across dimensionalities (2D and 3D) and modalities (natural and medical images) without requiring per-task tuning of its loss weight.

\section{Method}
\label{sec:method}

\begin{figure}[t]
\centering
\includegraphics[width=\linewidth]{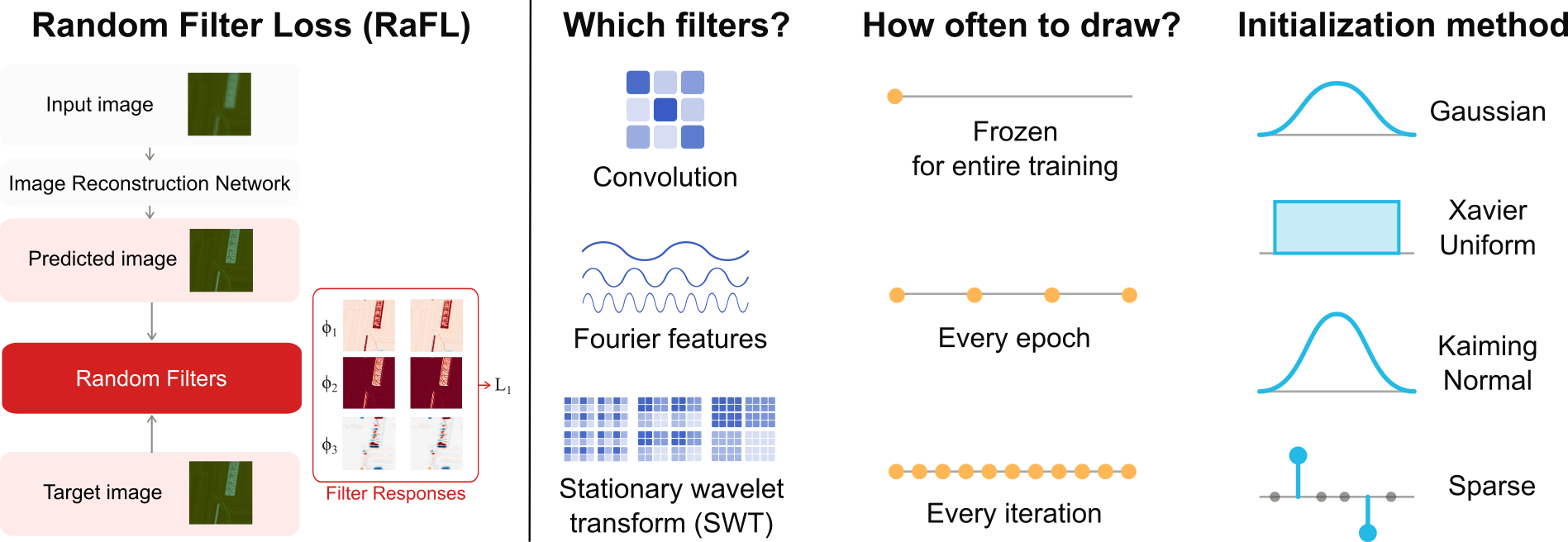}
\caption{\textbf{Overview of \method{}.} Left: training pipeline. Right: the three design choices studied for \method{}: filter family, weight redraw schedule, and initialization.}
\label{fig:method}
\end{figure}

In this work, we propose \method{}, a general random filter loss for both 2D and 3D image reconstruction that requires no pretrained weights, modality specific components, or no per-task filter selection.
\method{} is motivated by the idea that a signal can be probed through several different measurement filters applied to the same data~\citep{candes2006stable}.
Accordingly, we implement the loss with three filter families, each capturing a different aspect of the image. We characterize \method{} through an empirical study of three factors: (i) choice of filters, (ii) weight resampling schedule, and (iii) weight initialization method. This analysis clarifies which components drive performance and yields practical guidelines.
Using the resulting configuration, we then evaluate \method{} on 2D and 3D image reconstruction tasks.

\subsection{Choice of Filters}
\label{sec:bagdef}
The three families, a random convolution, random Fourier features and a random wavelet detail, trade off spatial and frequency resolution differently. Random convolutions are spatially local but broad in frequency, Fourier features compare small patches through random frequencies, and the random wavelet detail sits between them with joint space-frequency localization across scales. Combining them lets the loss penalize errors that any single family would miss, such as fine local texture, patch-level structure, or scale-specific detail. We empirically show this using image denoising on SIDD~\citep{abdelhamed2018high}.

\paragraph{Random convolution.}
Convolutional filters respond to local spatial patterns such as edges and textures. We use a random kernel as a feature extractor for the loss without any training, and following \citet{liu2021generic} a single kernel whose weights are drawn from a fixed distribution at a fixed size (e.g., $3{\times}3$, $5{\times}5$, $7{\times}7$), spanning all input channels, with no nonlinearity.
A single kernel size, however, operates at one spatial scale and responds broadly across frequencies.
Convolution is therefore spatially specific but frequency-unselective, in contrast to the other filters we introduce next.

\paragraph{Random Fourier features.}

We adopt the idea of random Fourier feature construction of \citep{rahimi2007random} and use random sinusoidal projections to measure discrepancies between prediction and target patches.
For each patch, $D$ random Fourier features were computed, $\sqrt{2/D}\cos(w^\top p+b)$, where the frequency $w\sim\mathcal{N}(0,2\gamma I)$ is a random vector and $b$ is a random phase, both redrawn at every randomization step and shared by prediction $p$ and target $q$.
The Fourier term uses the $\ell_1$ distance between the two feature vectors (Eq.~\eqref{eq:loss}). Tables~\ref{tab:gamma} and~\ref{tab:width} report ablations of $\gamma$ and $D$, respectively.

\paragraph{Random wavelet detail.}
The stationary wavelet transform (SWT) applies a wavelet filter bank without downsampling, decomposing the input into subbands that are localized in both space and frequency, at multiple scales and orientations.

Short daubechies wavelets have been shown to help in image tasks: in CNN image classification, db2 improve accuracy, while the longer may reduce it \citep{li2020wavelet}.
Considering this and the training patch sizes, we used Daubechies order 2 (db2).
At each of the 4 levels, an independent random unit vector is sampled. Ablation can be found in Table~\ref{tab:subband} and implementation details can be found in the supplementary materials (App.~\ref{sec:instantiation}).

\subsection{Randomization and Optimization}
\label{sec:optim}
\paragraph{Weight resampling schedule.} We compare three randomization schedules: one draw for the whole run, one per epoch and one per step. Because gradient-norm matching updates the loss weights during training, a per-step redraw introduces a time-varying objective. We therefore use MAdam~\citep{liu2026madam}, which is designed for optimization under changing loss weights (\S\ref{sec:objective}). The effect of the preconditioner is evaluated separately in App.~\ref{sec:madam}.

\paragraph{Weight initialization.}
\label{sec:init}
Due to the randomness of untrained networks, the gradient can explode or vanish \citep{liu2021generic}.
We compared four weight initializations for \method{}: Gaussian, with $\sigma=0.5$; Kaiming normal \citep{he2015delving}; Xavier uniform \citep{glorot2010understanding}; and sparse initialization \citep{martens2010deep}.

\subsection{Loss Function}
\label{sec:objective}
Let $\hat y=f_\theta(x)$ be the prediction of the reconstruction network $f_\theta$ for a degraded input $x$, and $y$ the target. Let $\phi_1$, $\phi_2$ and $\phi_3$ be the three families of \S\ref{sec:bagdef}, the convolution, the Fourier features and the wavelet detail.
Our objective is
\begin{equation}
\mathcal{L}(\hat y, y) \;=\; \sum_{j}\lambda_j\,\big\|\phi_j(\hat y)-\phi_j(y)\big\|_1, \qquad
\label{eq:loss}
\end{equation}
the $\ell_1$ distance between the filtered prediction and the filtered target, summed over the choice of filters. Each $\phi_j$ carries its own random weights, drawn at every randomization step as described in \S\ref{sec:bagdef}, and the same draw is applied to $\hat y$ and to $y$ and then discarded.

Unlike conventional multi-term losses, \method{} requires no manual tuning of loss weights. Every 50 steps, we adaptively set $\lambda_j$ so that $\|\lambda_j\nabla_{\hat y}\mathcal{L}_j\|$ is equal across the three filter families, where $\mathcal{L}_j=\|\phi_j(\hat y)-\phi_j(y)\|_1$ and $\nabla_{\hat y}$ is the gradient with respect to the network output, following gradient-norm matching~\citep{chen2018gradnorm}. This automatically balances the loss terms and removes the need for task-specific $\lambda$ selection or grid search. Because the loss weights change during training, Adam's running statistics may become inconsistent with the current objective. MAdam~\citep{liu2026madam} fixes this by keeping Fisher estimates for every pair of loss terms, combines them with the current $\lambda_j$ into a diagonal metric $\hat M$, and passes $\hat M^{-1}d$ to Adam~\citep{kingma2014adam} in place of the summed gradient $d$ (App.~\ref{sec:madam}).
\section{Results}
\label{sec:results}

\subsection{Experimental Settings}
\label{sec:exp_settings}
To validate \method{}, we adopt the same tasks and datasets from baseline papers\citep{zhou2023training,pfaehler2025untrained} with their best architectures.
For \textbf{2D image denoising}, the real-world Smartphone Image Denoising Dataset (SIDD) \citep{abdelhamed2018high}, paired with DnCNN \citep{zhang2017beyond} was selected.
In \textbf{2D image enhancement}, we adapted low-light (LOL) dataset~\citep{wei2018deep} with DRBN~\citep{yang2020fidelity}.
For \textbf{2D super-resolution}, pan-sharpening on WorldView-2 with SFINet \citep{zhou2022spatial} was used.
Following the \textbf{3D image denoising} setup for untrained perceptual loss (uPL) in \citet{pfaehler2025untrained}, we performed MR angiography denoising on the IXI dataset \citep{jiang2018denoising} at 5\% Rician noise \citep{gudbjartsson1995rician}, using a 3D DnCNN.
Since uPL was designed for 3D MR denoising, we additionally experimented with \textbf{3D image super-resolution} using T1-weighted brain MR images from OASIS-3 \citep{lamontagne2019oasis}. Images were $4\times$ downsampled and SwinUNETR \citep{hatamizadeh2021swin} was trained to recover the isotropic image.
\paragraph{Baseline Losses.} Four prototype losses from \citet{zhou2023training}, CDC, reverse, an INN and Taylor, are used for 2D baselines.
The pixel loss ($\ell_1$) and the uPL of \citet{pfaehler2025untrained}, which compares the features of an untrained CNN and has no pixel loss term, are used for both 2D and 3D baselines. We run uPL with the SimpleNet which paper recommends, on all the datasets so every task compare aginst the same. Also we run an extra untrained VGG19 version of uPL in SIDD dataset, which matches the architecture of the pretrained VGG as a comprehensive compare.
Within each task, methods share the backbone, data pipeline, training budget, and seeds ($42$--$49$), and are evaluated at the final epoch. The single-term baselines (pixel $\ell_1$ and both uPL variants) use standard Adam with unit loss weight. The four prototype baselines use Adam with fixed equal weights on the pixel and prototype terms, with the prototype redrawn once per epoch. We also evaluate pixel $\ell_1$ plus ImageNet-pretrained VGG19 with Adam on SIDD and LOL. \method{} updates the loss weights every $50$ steps by gradient-norm matching and uses the MAdam preconditioner~\citep{liu2026madam} followed by Adam. Further details are in App.~\ref{sec:instantiation}.

\paragraph{Implementation Details.}
Across all tasks, the initial learning rate is set to $10^{-3}$, except image enhancement on LOL, which uses $10^{-4}$. Datasets are strictly partitioned into disjoint training, validation, and test splits (approximately $8:1:1$) at the subject or scene level to prevent data leakage.
We selected the default configuration on the SIDD validation split, using mean PSNR as the primary criterion and cross-seed PSNR standard deviation as the secondary criterion.
Details about image preprocessing are in App.~\ref{sec:instantiation} and hyperparameters and further configuration are in App.~\ref{sec:ablations-more}.

\subsection{Ablation study answering three design questions}
\label{sec:threeq}\label{sec:bag}\label{sec:schedule}
We conduct extensive ablation studies on the SIDD image denoising task using DnCNN to systematically address three fundamental design questions for \method{}: which filters to use, how frequently to redraw them, and how to initialize their weights. Results are summarized in Table~\ref{tab:threeq}.

\begin{table}[t]
\centering\scriptsize
\setlength{\tabcolsep}{3pt}
\begin{tabular}{llcccc}
\toprule
& Configuration & PSNR $\uparrow$ & SSIM $\uparrow$ & LPIPS $\downarrow$ & DISTS $\downarrow$ \\
\midrule
& \textbf{\method{}}: all three, every step, Gaussian & $\mathbf{40.28 \pm 0.13}$ & $\mathbf{0.9557 \pm 0.0012}$ & $0.2488 \pm 0.0036$ & $0.1847 \pm 0.0034$ \\
\midrule
\multirow{7}{*}{\rotatebox{90}{Q1, which filters}}
& Wavelet only            & $32.26 \pm 3.13$ & $0.8872 \pm 0.0659$ & $0.2874 \pm 0.0240$ & $0.2077 \pm 0.0166$ \\
& Convolution only        & $39.75 \pm 0.17$ & $0.9523 \pm 0.0014$ & $0.2605 \pm 0.0045$ & $0.1907 \pm 0.0035$ \\
& Fourier only            & $39.92 \pm 0.32$ & $0.9548 \pm 0.0008$ & $0.2547 \pm 0.0047$ & $0.1878 \pm 0.0041$ \\
& Convolution + Wavelet   & $39.88 \pm 0.75$ & $0.9538 \pm 0.0015$ & $\mathbf{0.2446 \pm 0.0041}$ & $\mathbf{0.1825 \pm 0.0043}$ \\
& Convolution + Fourier   & \underline{$39.98 \pm 0.33$} & $0.9552 \pm 0.0010$ & $0.2573 \pm 0.0061$ & $0.1907 \pm 0.0052$ \\
& Fourier + Wavelet       & $39.28 \pm 1.36$ & $0.9548 \pm 0.0010$ & $0.2498 \pm 0.0052$ & $0.1894 \pm 0.0081$ \\
\cmidrule{2-6}
& pixel $\ell_1$          & $39.64 \pm 1.00$ & $0.9551 \pm 0.0011$ & $0.2478 \pm 0.0056$ & $0.1830 \pm 0.0057$ \\
\midrule
\multirow{2}{*}{\rotatebox{90}{Q2}}
& Every epoch     & $39.78 \pm 0.67$ & $0.9545 \pm 0.0014$ & $0.2503 \pm 0.0054$ & $0.1863 \pm 0.0068$ \\
& Fixed for the entire run   & $39.67 \pm 1.03$ & $0.9546 \pm 0.0011$ & \emph{$0.2494 \pm 0.0052$} & $0.1852 \pm 0.0056$ \\
\midrule
\multirow{3}{*}{\rotatebox{90}{Q3}}
& Kaiming normal          & $39.92 \pm 0.60$ & \underline{$0.9552 \pm 0.0008$} & $0.2508 \pm 0.0064$ & $0.1867 \pm 0.0059$ \\
& Xavier uniform          & \underline{$39.98 \pm 0.41$} & $0.9548 \pm 0.0016$ & $0.2529 \pm 0.0039$ & $0.1876 \pm 0.0042$ \\
& Sparse                  & $39.91 \pm 0.31$ & $0.9542 \pm 0.0011$ & \underline{$0.2468 \pm 0.0033$} & \underline{$0.1830 \pm 0.0040$} \\
\bottomrule
\end{tabular}

\caption{\textbf{Ablation study of filter family (Q1), redraw schedule (Q2), and initialization (Q3) on SIDD.} (DnCNN, eight seeds per configuration; mean ± s.d. across seeds; bold indicates the best and underline the second-best value in each column) \method{} uses all three filter families, with filter weights redrawn from a Gaussian distribution at every optimization step (first row).
}
\label{tab:threeq}\label{tab:sidd}
\end{table}

\paragraph{Choice of filters}
\method{} with all three filter families, weights redrawn every iteration in Gaussian initialization, achieves the highest PSNR ($40.28$\,dB) and SSIM ($0.9557$), outperforming every proper subset. The Convolution+Wavelet pair achieves the best perceptual quality, LPIPS (0.2446) and DISTS (0.1825). Sparse initialization with all three families is a close second on both (LPIPS 0.2468, DISTS 0.1830), and still surpasses \method{}'s own Gaussian-initialized result on these two metrics. Across initialization methods (Table~\ref{tab:threeq}, Q3), no configuration collapses on any metric, unlike using a partial set of filter families, such as Wavelet only on PSNR. In a frequency analysis of the SIDD test error (App.~\ref{sec:fourier-bands}), removing the Fourier family (Convolution + Wavelet) lowers mean PSNR across all three frequency bands, with the largest drop in the high-frequency band.
We further evaluate convolution kernel size, additional convolution filters, and with and without handcrafted filters including Sobel and Laplacian (Table~\ref{tab:ladder}).
A $7^2$ convolution achieves the highest PSNR and SSIM with the smallest seed variance, while a $3^2$ convolution is lower by $0.23$\,dB PSNR on seven of eight seeds. Adding additional convolution scales, Sobel filters, or Laplacian filters does not yield consistent improvements, suggesting that the three filter families are sufficient for the tasks studied.
We also train a standalone deterministic composite, pixel $\ell_1$ + fixed Sobel + fixed Laplacian, on SIDD with DnCNN over the same eight seeds (Table~\ref{tab:ladder}). \method{} outperforms this composite on all four metrics, improving mean PSNR by $0.95$\,dB with higher SSIM and lower LPIPS and DISTS.

\paragraph{Weight resampling schedule.}
Redrawing every step outperforms both draw held for an epoch and a draw held for the entire run on all four metrics (Table~\ref{tab:threeq} Q2). Compared to holding weights fixed per epoch or for the entire training run, per-step randomization improves PSNR by $0.50$\,dB and $0.61$\,dB, respectively.
Per-step redrawing significantly reduces performance variance across random seeds ($\pm 0.13$\,dB vs. $\pm 0.67$\,dB and $\pm 1.03$\,dB), preventing optimization from getting trapped into suboptimal configurations.
We further evaluate this redraw schedule across all five tasks in App.~Table~\ref{tab:sched_full}. Per-step resampling achieves the highest mean PSNR and overall performance across the five tasks. Additional analyses of loss weighting and preconditioning are reported in App.~Tables~\ref{tab:lambda} and \ref{tab:madam}.

\paragraph{Weight initialization.}
Weight initialization has comparatively little influence on performance (Table~\ref{tab:threeq}). The four initialization strategies differ by at most $0.38$\,dB in PSNR, while the choice of filter families changes performance by as much as $8.02$\,dB. Although prior work has highlighted initialization as an important consideration for deep random networks~\citep{liu2021generic}, our findings are more consistent with~\citet{pfaehler2025untrained}, who observed only small differences across initialization schemes. We use Gaussian initialization for its highest mean PSNR and smallest cross-seed PSNR spread on SIDD. On WorldView-2, the four initializations are within $0.05$\,dB in mean PSNR (App.~\ref{sec:transfer-ablations}).

\subsection{Experimental Results on Image Reconstruction Tasks}
\label{sec:apply}

\paragraph{2D tasks: image denoising, enhancement, and super-resolution.}
Across the three 2D tasks, \method{} matches or improves upon the pixel loss and is competitive with or better than prior random filter losses (Table~\ref{tab:apply2d}). Notably, the same configuration selected in Sec.~\ref{sec:threeq} is transferred unchanged across all tasks.

\begin{figure}[t]
\centering
\includegraphics[width=0.9\linewidth]{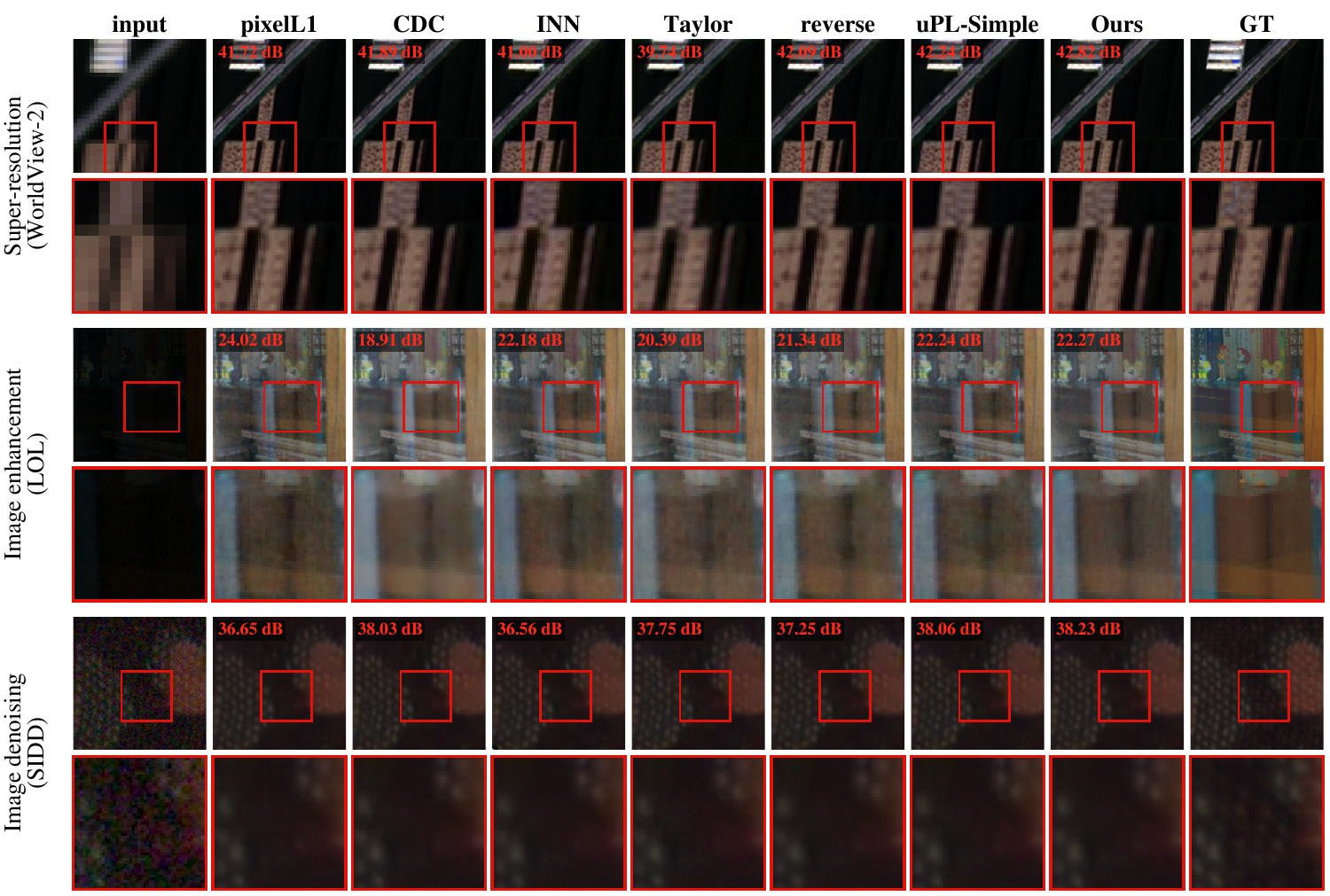}
\caption{\textbf{Representative examples for 2D image reconstruction tasks.} Columns compare competing baseline losses (pixel loss, four prototypes from \citet{zhou2023training}, uPL \citet{pfaehler2025untrained}, \method{} (Ours), and the ground truth. PSNR is shown on each panel.}\label{fig:qual2d}
\end{figure}

On SIDD, \method{} achieves the highest mean PSNR and is statistically indistinguishable from the strongest baseline, CDC ($40.28$ vs.\ $40.26$\,dB; paired interval $[-0.12,+0.15]$). Both methods outperform the pixel loss by more than $0.6$\,dB, whereas the perceptual metrics differ only within the observed seed variability. Qualitative examples (Fig.~\ref{fig:qual2d}) show that \method{} preserves fine image texture while avoiding the oversmoothing visible in several competing methods.

On LOL, PSNR differences among \method{}, the pixel loss, and uPL are small ($21.17$, $21.16$, and $21.33$\,dB). Among methods without pretrained features, \method{} achieves the lowest LPIPS and DISTS. Fig.~\ref{fig:qual2d} (middle) shows that \method{} better preserves the wood grain texture and local contrast, whereas several competing methods introduce color artifacts and blur fine structural details.

On SIDD and LOL, pixel $\ell_1$ with pretrained VGG19 achieves lower LPIPS and DISTS than \method{}, as well as higher SSIM on LOL, but yields lower mean PSNR by $1.00$\,dB on SIDD and $0.43$\,dB on LOL. Pretrained VGG19 features are designed for 2D RGB images and require adaptation to multispectral and 3D inputs, whereas \method{} can be applied directly to these modalities.

The largest improvement is observed on WorldView-2, where \method{} outperforms the pixel loss by $1.51$\,dB and the strongest prototype (CDC) by $0.87$\,dB. Furthermore, \method{} achieves the best value on all four evaluation metrics. While uPL remains competitive, it falls behind \method{} across every metric. Together, these results indicate that the proposed filter combination is particularly beneficial for reconstruction tasks requiring recovery of high-frequency spatial and spectral information.

\begin{table}[t]
\centering\small
\setlength{\tabcolsep}{5pt}
\begin{tabular}{lcccc}
\toprule
\multicolumn{5}{l}{\emph{Image denoising, SIDD}} \\
Loss & PSNR $\uparrow$ & SSIM $\uparrow$ & LPIPS $\downarrow$ & DISTS $\downarrow$ \\
\midrule
\textbf{\method{}} & $\mathbf{40.28 \pm 0.13}$ & \underline{$0.9557 \pm 0.0012$} & $0.2488 \pm 0.0036$ & $0.1847 \pm 0.0034$ \\
pixel $\ell_1$ & $39.64 \pm 1.00$ & $0.9551 \pm 0.0011$ & \underline{$0.2478 \pm 0.0056$} & \underline{$0.1830 \pm 0.0057$} \\
pixel $\ell_1$ + pretrained VGG19 & $39.28 \pm 0.26$ & $0.9443 \pm 0.0014$ & $\mathbf{0.2328 \pm 0.0027}$ & $\mathbf{0.1807 \pm 0.0033}$ \\
CDC     & \underline{$40.26 \pm 0.06$} & $\mathbf{0.9568 \pm 0.0004}^{*}$& $0.2681 \pm 0.0057$ & $0.2012 \pm 0.0049$ \\
reverse & $39.33 \pm 0.71$ & $0.9550 \pm 0.0015$ & $0.2560 \pm 0.0061$ & $0.1905 \pm 0.0054$ \\
Taylor  & $38.20 \pm 1.87$ & $0.9523 \pm 0.0021$ & $0.2655 \pm 0.0118$ & $0.2027 \pm 0.0199$ \\
INN     & $33.92 \pm 5.21$ & $0.9447 \pm 0.0194$ & $0.2822 \pm 0.0329$ & $0.2338 \pm 0.0386$ \\
uPL, SimpleNet & $40.09 \pm 0.07$ & $0.9541 \pm 0.0005$ & $0.2648 \pm 0.0021$ & $0.1957 \pm 0.0026$ \\
uPL, VGG19     & $39.18 \pm 0.16$ & $0.9410 \pm 0.0020$ & $0.2631 \pm 0.0058$ & $0.1903 \pm 0.0032$ \\
\midrule
\multicolumn{5}{l}{\emph{Image enhancement, LOL}} \\
Loss & PSNR $\uparrow$ & SSIM $\uparrow$ & LPIPS $\downarrow$ & DISTS $\downarrow$ \\
\midrule
\textbf{\method{}} & \underline{$21.17 \pm 0.33$} & $0.8035 \pm 0.0032$ & \underline{$0.2568 \pm 0.0043$} & \underline{$0.1995 \pm 0.0032$} \\
pixel $\ell_1$ & $21.16 \pm 0.46$ & $0.7799 \pm 0.0031$ & $0.3055 \pm 0.0075$ & $0.2273 \pm 0.0049$ \\
pixel $\ell_1$ + pretrained VGG19 & $20.74 \pm 0.59$ & $\mathbf{0.8292 \pm 0.0051}$ & $\mathbf{0.1946 \pm 0.0038}$ & $\mathbf{0.1655 \pm 0.0020}$ \\
CDC     & $20.55 \pm 0.31$ & \underline{$0.8046 \pm 0.0047$} & $0.2863 \pm 0.0061$ & $0.2141 \pm 0.0052$ \\
reverse & $21.08 \pm 0.38$ & $0.7998 \pm 0.0018$ & $0.2827 \pm 0.0036$ & $0.2146 \pm 0.0028$ \\
Taylor  & $20.58 \pm 0.32$ & $0.7893 \pm 0.0038$ & $0.2904 \pm 0.0084$ & $0.2199 \pm 0.0041$ \\
INN     & $20.74 \pm 0.35$ & $0.8000 \pm 0.0053$ & $0.2804 \pm 0.0065$ & $0.2129 \pm 0.0045$ \\
uPL     & $\mathbf{21.33 \pm 0.43}$ & $0.7880 \pm 0.0111$ & $0.2951 \pm 0.0169$ & $0.2220 \pm 0.0094$ \\
\midrule
\multicolumn{5}{l}{\emph{Super-resolution, WorldView-2}} \\
Loss & PSNR $\uparrow$ & SSIM $\uparrow$ & ERGAS $\downarrow$ & SAM $\downarrow$ \\
\midrule
\textbf{\method{}} & $\mathbf{39.62 \pm 0.08}^{*}$& $\mathbf{0.9631 \pm 0.0005}^{*}$& $\mathbf{1.318 \pm 0.011}^{*}$& $\mathbf{0.0297 \pm 0.0002}^{*}$\\
pixel $\ell_1$ & $38.10 \pm 0.13$ & $0.9531 \pm 0.0011$ & $1.532 \pm 0.026$ & $0.0349 \pm 0.0005$ \\
CDC     & $38.75 \pm 0.14$ & $0.9594 \pm 0.0011$ & $1.437 \pm 0.024$ & $0.0326 \pm 0.0005$ \\
reverse & $38.70 \pm 0.12$ & $0.9591 \pm 0.0009$ & $1.434 \pm 0.020$ & $0.0326 \pm 0.0005$ \\
Taylor  & $37.90 \pm 0.71$ & $0.9493 \pm 0.0086$ & $1.670 \pm 0.206$ & $0.0367 \pm 0.0030$ \\
INN     & $37.49 \pm 0.67$ & $0.9457 \pm 0.0076$ & $1.783 \pm 0.214$ & $0.0389 \pm 0.0033$ \\
uPL     & \underline{$39.15 \pm 0.20$} & \underline{$0.9603 \pm 0.0016$} & \underline{$1.401 \pm 0.040$} & \underline{$0.0314 \pm 0.0008$} \\
\bottomrule
\end{tabular}
\caption{\textbf{Results on 2D image reconstruction tasks.} uPL follows \citep{pfaehler2025untrained}; CDC, reverse, Taylor, and INN follow \citep{zhou2023training}. The symbol $^{*}$ denotes significance against the second-best method (paired t-test, $p<0.05$). Paired significance is reported for comparisons between methods without pretrained features. The pretrained VGG19 uses ImageNet-pretrained features. Values are mean $\pm$ s.d. across 8 seeds.}

\label{tab:apply2d}\label{tab:lol}\label{tab:pan}
\end{table}

\begin{table}[t]
\centering\small
\setlength{\tabcolsep}{5pt}
\begin{tabular}{lcccc}
\toprule
Loss & PSNR $\uparrow$ & SSIM $\uparrow$ & LPIPS $\downarrow$ & DISTS $\downarrow$ \\
\midrule
\multicolumn{5}{l}{\emph{Super-resolution, OASIS3}} \\
\textbf{\method{}} & $\mathbf{22.22 \pm 0.03}^{*}$& $\mathbf{0.8249 \pm 0.0017}$& $\mathbf{0.2962 \pm 0.0010}^{*}$& $\mathbf{0.2547 \pm 0.0010}$\\
 pixel $\ell_1$ & \underline{$21.89 \pm 0.06$} & \underline{$0.7782 \pm 0.0768$} & \underline{$0.3029 \pm 0.0060$} &\underline{$0.2592 \pm 0.0069$} \\
uPL & $18.48 \pm 0.24$ & $0.5294 \pm 0.0521$ & $0.4104 \pm 0.0169$ & $0.3158 \pm 0.0080$ \\
\midrule
\multicolumn{5}{l}{\emph{Image denoising, IXI}} \\
\textbf{\method{}} & $\mathbf{29.66 \pm 0.19}$ & \underline{$0.9627 \pm 0.0007$} & $0.0646 \pm 0.0012$ & \underline{$0.0785 \pm 0.0013$} \\
pixel $\ell_1$ & $29.56 \pm 0.33$ & $0.9622 \pm 0.0014$ & \underline{$0.0645 \pm 0.0021$} & $0.0788 \pm 0.0029$ \\
uPL & \underline{$29.63 \pm 0.35$} & $\mathbf{0.9628 \pm 0.0014}$ & $\mathbf{0.0632 \pm 0.0016}$ & $\mathbf{0.0775 \pm 0.0025}$ \\
\bottomrule
\end{tabular}
\caption{\textbf{Results on 3D image reconstruction tasks.} LPIPS and DISTS for super-resolution task are averaged over all slices of the three orthogonal planes. The symbol $^{*}$ denotes significance against the second-best method (paired t-test, $p<0.05$); mean $\pm$ s.d. across 8 seeds.}
\label{tab:apply3d}\label{tab:oasis3}\label{tab:ixi}
\end{table}

\begin{figure}[t]
\centering
\includegraphics[width=0.9\linewidth]{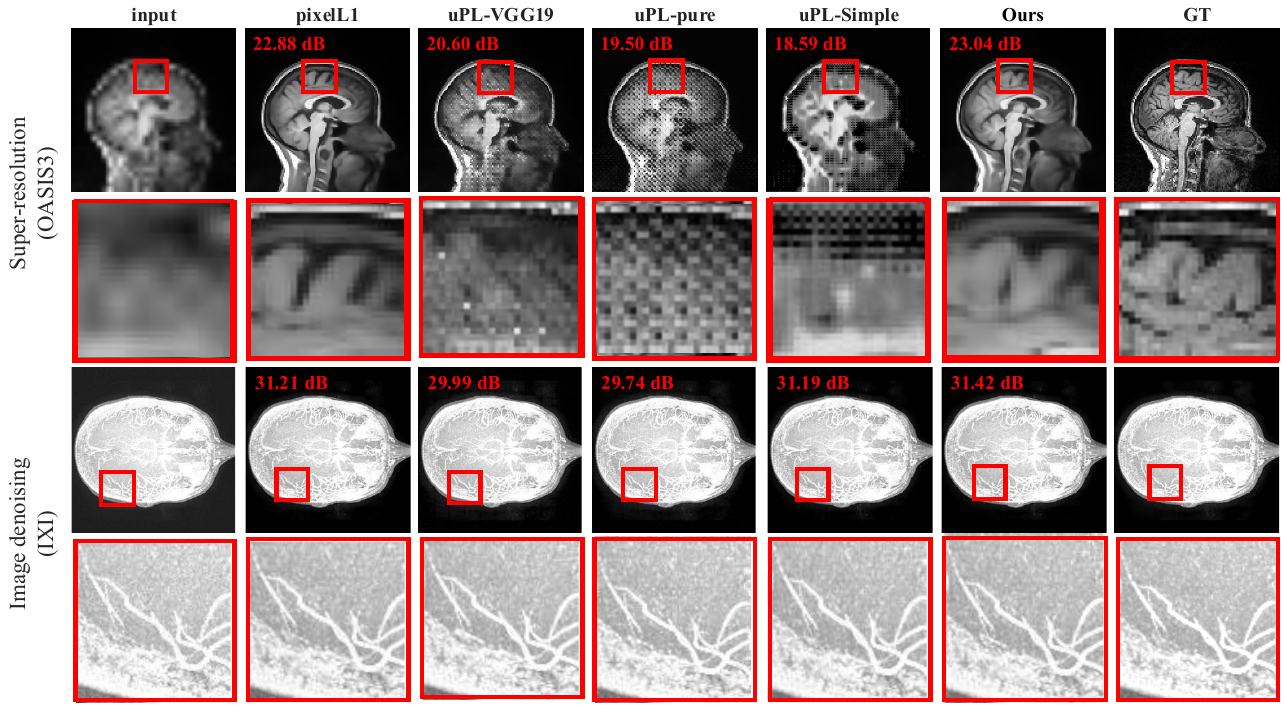}
\caption{\textbf{Representative examples for 3D image reconstruction tasks.}
Columns compare competing losses (from left to right): pixel loss, uPL-VGG19, uPL-pure (untrained VGG19), uPL-Simple (3D SimpleNet of \citep{pfaehler2025untrained}), and \method{}.PSNR is shown on each panel.}

\label{fig:qual3d}
\end{figure}

\paragraph{3D tasks: super-resolution on OASIS3 and denoising on IXI.}
On $4\times$ super-resolution of OASIS3, \method{} consistently outperforms both the pixel loss and uPL (Table~\ref{tab:apply3d}). It achieves the best value on all four metrics and the highest PSNR on every seed, significantly improving upon the pixel loss by $0.33$\,dB and uPL by $3.74$\,dB. The small variance across seeds ($\pm0.03$\,dB) further indicates stable optimization.
Representative examples in Fig.~\ref{fig:qual3d} show improved recovery of fine anatomical structures relative to both baselines.
The pixel loss gives the cerebellar folia as straight striping and lost sulcus overall, \method{} follows their curve as the ground truth does, and the untrained network introduces a regular lattice-like texture that is not present in the ground truth image.

On MR angiography denoising with a 3D DnCNN on IXI \citep{jiang2018denoising} at $5\%$ Rician noise \citep{gudbjartsson1995rician}, \method{}, the pixel loss, and uPL achieve similar performance (Table~\ref{tab:apply3d}). \method{} attains the highest mean PSNR ($29.66$\,dB), although the difference relative to both baselines falls within the observed seed variability. While uPL achieves the best mean SSIM, LPIPS, and DISTS, the differences among the three methods are small.

\paragraph{Optimizer.}

The adaptive weighting strategy is held fixed across all tasks by evaluated once. Table~\ref{tab:lambda} shows that gradient-norm matching consistently outperforms fixed uniform weighting, uncertainty weighting \citep{kendall2018multi}, and randomly sampled Dirichlet weights. On SIDD, gradient-norm matching improves PSNR by $0.23$--$0.77$\,dB relative to these alternatives, with every competing weighting strategy falling behind on all eight seeds. It needs each term's gradient with respect to the network output once every $50$ steps, which does not pass back through the network, and the weights it settled are listed in Table~\ref{tab:final_lambda}.
As shown in Table~\ref{tab:madam}, MAdam consistently improves or matches Adam alone.
The largest benefit is observed on WorldView-2, where PSNR increases by $0.87$\,dB, while LOL improves by $0.26$\,dB and SIDD remains statistically similar ($+0.09$\,dB).
These results suggest that accounting for changing objective weights becomes increasingly important on more challenging reconstruction tasks. Additionally, the weights learned by gradient-norm matching are stable across both seeds and tasks (App.~Table~\ref{tab:final_lambda}), with the Fourier family consistently receiving the largest weight and the convolution family the smallest.

Overall, \method{} either improves upon or matches the strongest baseline on all five tasks while using the same filter families, redraw schedule, and initialization, without task-specific selection of loss weights. This suggests that the resulting objective transfers across various reconstruction problems.

\subsection{Loss Complexity}
\label{sec:cost}\label{sec:balancebody}

We compare the computational cost of the loss functions (App.~Table~\ref{tab:cost}). On SIDD, the \method{} loss costs $1.12$ GFLOPs, corresponding to only $1.9\%$ of the DnCNN forward pass. This is substantially lower than both the CDC prototype ($5.00$ GFLOPs) and the VGG19-based perceptual loss ($203.8$ GFLOPs), while remaining competitive with filter based losses. On OASIS3, the relative overhead is even smaller, accounting for only $1.0\%$ of the SwinUNETR forward pass. In contrast, the VGG19 form of uPL requires $1821$ GFLOPs, corresponding to $60.7\%$ of the combined loss and backbone forward FLOPs.
Because the random filters are discarded after training, \method{} adds nothing at inference. FLOPs are measured using \texttt{FlopCounterMode} and implementation details are provided in App.~\ref{sec:instantiation}.
Wall-clock time does not scale directly with FLOPs, as the small random filter operations in \method{} are less hardware-efficient than large convolutional kernels (see \S\ref{sec:implementation} for details). However, this overhead is incurred only during training and does not affect inference. To further reduce it, we optimized the implementation (App.~\ref{apdx:lossComplexity}). With this implementation, a training step on SIDD takes $8.3$\,ms with \method{} ($4.6$\,ms with Adam alone), against $3.3$\,ms with pixel $\ell_1$. We will release this implementation as well.
\section{Discussion and conclusion}
\label{sec:discussion}
\label{sec:conclusion}
\label{sec:limits}
In this work, we investigated which design choices matter most in random-filter losses, focusing on three factors: filter family, weight resampling schedule, and initialization. Building on these findings, we introduced RaFL, a random-filter loss that we evaluated on five image reconstruction tasks spanning natural and medical images in both 2D and 3D.

The choice of filter families changes mean PSNR more than the redraw schedule or the initialization does. Based on validation PSNR and cross-seed stability, we adopt a default configuration that combines all three families with per-step Gaussian redraws (Sec.~\ref{sec:threeq}). Using the configuration selected on SIDD unchanged across all remaining tasks, RaFL improves upon pixel loss on denoising, pan-sharpening, and super-resolution, while remaining competitive on low-light enhancement and MR angiography denoising. The same filter families, redraw schedule, initialization, and adaptive weighting strategy were applied throughout, without manual loss-weight selection, demonstrating that a single random-filter objective can transfer across reconstruction problems spanning natural and medical images in both 2D and 3D.

Several limitations remain. Although \method{} adds little computational overhead in terms of FLOPs and no inference cost, per-step resampling and gradient balancing increase wall-clock training time. An optimized implementation partly mitigates this overhead (App.~Table~\ref{tab:cost}). In addition, while the same configuration transferred across all five tasks, the design study is conducted on two 2D datasets, SIDD and WorldView-2 (App.~\ref{sec:transfer-ablations}). Finally, understanding why the learned loss weights exhibit similar patterns across datasets and modalities remains an interesting direction for future work.

\section*{Use of AI Assistants}

In this work, we used generative AI tools (Claude Code) solely for aiding with language polishing, grammar correction in this manuscript. We did not use generative AI tools for any of the other tasks that ICLR requires authors to disclose, including generating data, developing theory, formulating or proving mathematical claims, proposing hypotheses, designing experiments, implementing core algorithmic methods, processing data, or interpreting results. All AI-assisted text revisions have been thoroughly reviewed and edited by the authors to ensure factual and technical accuracy. The authors take full responsibility for the final content, claims, and artifacts presented in this paper.
\section*{Reproducibility statement}
Every number in \S\ref{sec:results} is a per-image measurement on the same held-out split, over the seeds stated with each table, and the code and the evaluation files behind every table will be released. The data are public (SIDD, LOL, WorldView-2, OASIS3, IXI).

\S\ref{sec:results} gives the preprocessing, the split, the backbone, the optimizer and the weighting strategy. Every configuration is read at its last epoch rather than at a selected checkpoint, and App.~\ref{sec:instantiation} gives every filter of every configuration. The SIDD splits are scene disjoint and the design choices were made on the validation split, so no configuration in this paper was selected by reading the test split (\S\ref{sec:results}).

The three families are not normalized alike: the wavelet draws have unit norm and the convolution and Fourier draws do not. A constant on any one term is therefore a reweighting of it, which is why the loss weights are set by gradient-norm matching and why the weighting, the preconditioner and the optimizer are held identical across every configuration outside Tables~\ref{tab:lambda} and \ref{tab:madam}, which vary them.

\bibliographystyle{iclr2027_conference}
\bibliography{main}

\ifwithsupp
\clearpage
\appendix
\renewcommand{\thetable}{S\arabic{table}}\setcounter{table}{0}
\renewcommand{\thefigure}{S\arabic{figure}}\setcounter{figure}{0}
\renewcommand{\theequation}{S\arabic{equation}}\setcounter{equation}{0}
\begin{center}\Large\bfseries Supplementary Material\end{center}
\suppressfloats[t]
\appendix
\setcounter{table}{0}\renewcommand{\thetable}{S\arabic{table}}
\setcounter{figure}{0}\renewcommand{\thefigure}{S\arabic{figure}}
\renewcommand{\theHtable}{S\arabic{table}}\renewcommand{\theHfigure}{S\arabic{figure}}

\section{Additional Experimental Results}
\label{sec:bodytables}\label{sec:ablations-more}

\subsection{Random Filter Families}
\label{sec:families}
Table~\ref{tab:bagdef} lists the random filter implementations. The first group in Table~\ref{tab:ladder} is the three families of \method{}:  random convolution with a $7\times7$ kernel ($7^3$ in 3D), random Fourier features and a random wavelet detail. The second group contains the four additional filters that appear only in the ablation of Table~\ref{tab:ladder}: random convolutions with kernel size $3$ and $5$, random-orientation Sobel and random Laplacian filter. Every filter is redrawn at every optimization step, applied to the prediction and the target with the same draw, and discarded afterwards. The convolutions draw every weight from $\mathcal{N}(0,0.5^2)$, span all input channels, and apply no nonlinearity or rescaling. The Sobel filter points in a random direction and the Laplacian takes a random symmetric form. The random Fourier features and the random wavelet detail are described in the method section.

Table~\ref{tab:ladder} varies the kernel of the convolution and adds the extra filters. Among the settings, conv $7$ has the highest PSNR and SSIM and the lowest standard deviation across seed. It is ahead of $5$ and $3$ by $0.29$ and $0.23$\,dB, separately. The support sets the neighborhood over which the loss compares the error, so a larger kernel lets the loss has more receptive field of its neighbors. Adding more filters does not help: using the convolution at all three(five filters) supports lowers PSNR by $0.31$\,dB, adding the Sobel and Laplacian filters(seven filters) lowers it further, and removing the convolutions from that seven-filter set leaves it behind \method{} on PSNR, LPIPS and DISTS. Therefore we keep the three families with a single $7\times7$ convolution.

\begin{table}[h!]
\centering\scriptsize
\setlength{\tabcolsep}{4pt}
\begin{tabular}{lcc@{\ }cc}
\toprule
Filters in the loss & PSNR $\uparrow$ & SSIM $\uparrow$& LPIPS $\downarrow$& DISTS $\downarrow$\\
\midrule
\textbf{all three, conv $7^2$ (\method{})} & $\mathbf{40.28 \pm 0.13}$ & $0.9557 \pm 0.0012$& $\underline{0.2488 \pm 0.0036}$& $\underline{0.1847 \pm 0.0034}$\\
all three, conv $5^2$    & $39.99 \pm 0.73$ & $0.9547 \pm 0.0009$& $\mathbf{0.2464 \pm 0.0034}$& $\mathbf{0.1840 \pm 0.0067}$\\
all three, conv $3^2$    & $40.06 \pm 0.29$ & $0.9553 \pm 0.0011$& $0.2522 \pm 0.0052$& $0.1882 \pm 0.0043$\\
\midrule
five filters             & $39.98 \pm 0.42$ & $0.9552 \pm 0.0015$& $0.2517 \pm 0.0050$& $0.1870 \pm 0.0048$\\
seven filters            & $39.75 \pm 0.76$ & $\mathbf{0.9562 \pm 0.0009}$& $0.2602 \pm 0.0054$& $0.1946 \pm 0.0054$\\
seven minus the convolutions & $\underline{40.06 \pm 0.21}$& $\underline{0.9559 \pm 0.0010}$& $0.2633 \pm 0.0090$& $0.1968 \pm 0.0063$\\
\midrule
pixel $\ell_1$ + fixed Sobel + fixed Laplacian & $39.33 \pm 0.73$ & $0.9544 \pm 0.0009$ & $0.2547 \pm 0.0055$ & $0.1875 \pm 0.0057$ \\
\bottomrule
\end{tabular}
\caption{\textbf{Three families at three convolution supports, and more than three filters} (SIDD, DnCNN, eight seeds, mean $\pm$ s.d.\ across seeds). The first block changes only the support of the random convolution; the second adds filters beyond the three families: the five-filter bag(\method{} + Conv5 + Conv3) uses the convolution at all three supports, and the seven-filter bag (five-filter bag +  Sobel + Laplacian) adds a random Sobel and a random Laplacian. The last row is a standalone pixel loss with fixed Sobel and Laplacian terms. Bold is the best value in each column and underlined the second best.}
\label{tab:ladder}
\end{table}

\begin{table}[H]
\centering\scriptsize
\setlength{\tabcolsep}{3pt}
\begin{tabular}{p{2.0cm}p{3.5cm}p{5.2cm}p{2.1cm}}
\toprule
Member & What is drawn & Constraint on the draw & Support \\
\midrule
random conv & kernel weights, i.i.d.\ $\mathcal{N}(0,0.5^2)$ & none: no nonlinearity, no normalization & $3^3$ \\
random conv & kernel weights, i.i.d.\ $\mathcal{N}(0,0.5^2)$ & none: no nonlinearity, no normalization & $5^3$ \\
random conv & kernel weights, i.i.d.\ $\mathcal{N}(0,0.5^2)$ & none: no nonlinearity, no normalization & $7^3$ \\
random wavelet & direction $u\sim\mathcal{U}(S^2)$ at each of the four levels & db2 stationary detail subbands, one shared cascade; zero at DC; unit norm & $3(2^s{-}1){+}1$ per axis: $4,10,22,46$ \\
random Laplacian & symmetric $A$, six Gaussian entries, $\|A\|_F{=}1$; eight per step & $\sum_a A_{aa}\partial_{aa}+2\sum_{a<b}A_{ab}\partial_{ab}$; kills constants and ramps & $3^3$ \\
random Sobel & direction $u\sim\mathcal{U}(S^2)$; eight per step & $\sum_a u_aS_a$, $S_a=[-1,0,1]\otimes[1,2,1]\otimes[1,2,1]$ along $a$; not normalized; depthwise & $3^3$ \\
random Fourier & $W\sim\mathcal{N}(0,2\gamma I)$, $b\sim\mathcal{U}[0,2\pi)$; $\gamma{=}1$, $D{=}64$ & $\phi(x)=\sqrt{2/D}\cos(W^\top x+b)$; nonlinear; feature discrepancy measured with $\ell_1$ & $3^3$ patch, $27$-dim \\
\bottomrule
\end{tabular}
\caption{\textbf{Every random filter used in this paper.} The first three rows are the random convolution at its three supports, of which $7^3$ is the one \method{} deploys alongside the Fourier and wavelet rows, the $3^3$ and $5^3$ convolutions, the Laplacian and the Sobel are the extra members of the seven-filter row of Table~\ref{tab:ladder}. The wavelet, Sobel and Laplacian are zero at DC by construction; the convolutions are unconstrained and their expected response is flat across the band.}
\label{tab:bagdef}
\end{table}

\subsection{Weighting the Filter Loss Terms}
\label{sec:balance}

Training uses two separate components. \emph{Gradient-norm matching} sets the loss weights $\lambda_j$ of Eq.~\eqref{eq:loss}, and the \emph{MAdam preconditioner} (App.~\ref{sec:madam}) transforms the weighted gradient before the Adam step. The weights are needed because the three terms differ in scale. Every $50$ steps, gradient-norm matching computes the gradient of each term $\mathcal{L}_j$ with respect to the network output $\hat y$ on its own and resets each $\lambda_j$ so that the three weighted gradients $\lambda_j\nabla_{\hat y}\mathcal{L}_j$ have the same norm, a closed-form variant of GradNorm~\citep{chen2018gradnorm}. The weights stay fixed between resets, and a reset does not pass back through the network.

\begin{table}[t]
\centering\scriptsize
\setlength{\tabcolsep}{4pt}
\begin{tabular}{lcccc}
\toprule
Weighting of the three terms & PSNR $\uparrow$ & SSIM $\uparrow$ & LPIPS $\downarrow$ & DISTS $\downarrow$ \\
\midrule
\textbf{gradient-norm matching (Ours)} & $\mathbf{40.28 \pm 0.13}$ & $\mathbf{0.9557 \pm 0.0012}$ & $\mathbf{0.2488 \pm 0.0036}$ & $\mathbf{0.1847 \pm 0.0034}$ \\
Dirichlet, redrawn every step \citep{lin2021reasonable} & $\underline{40.05 \pm 0.14}$ & $0.9540 \pm 0.0010$ & $0.2565 \pm 0.0037$ & $0.1894 \pm 0.0038$ \\
fixed uniform & $39.92 \pm 0.26$ & $0.9545 \pm 0.0010$ & $0.2551 \pm 0.0062$ & $\underline{0.1887 \pm 0.0049}$ \\
learned uncertainty \citep{kendall2018multi} & $39.51 \pm 0.85$ & $\underline{0.9548 \pm 0.0010}$ & $\underline{0.2535 \pm 0.0064}$ & $0.1905 \pm 0.0076$ \\
\bottomrule
\end{tabular}
\caption{\textbf{How the three terms are weighted} (SIDD, DnCNN, eight seeds per configuration, mean $\pm$ s.d.\ across seeds). The three families, the MAdam preconditioner and Adam are held fixed and only the rule for $\lambda_j$ changes, so the first row is the deployed configuration of Table~\ref{tab:threeq}. Bold is the best value in each column and underlined the second best.}
\label{tab:lambda}
\end{table}

We compare four schemes for weighting $\lambda_j$ on SIDD at eight seeds, with the three families, the MAdam preconditioner and Adam held fixed. The four are fixed uniform weights, weights redrawn from a Dirichlet at every step \citep{lin2021reasonable}, learned uncertainty weights \citep{kendall2018multi} and gradient-norm matching. Table~\ref{tab:lambda} shows that gradient-norm matching is the best weighting: it has the highest mean PSNR ($40.28\pm0.13$\,dB) and the lowest standard deviation across seed of the four, while the spreads of the other three reach $\pm0.85$\,dB.

\begin{table}[t]
\centering\scriptsize
\setlength{\tabcolsep}{4pt}
\begin{tabular}{lccc}
\toprule
Task & Convolution & Wavelet & Fourier \\
\midrule
Denoising, SIDD & $0.046 \pm 0.001$ & $0.201 \pm 0.003$ & $0.754 \pm 0.003$ \\
Enhancement, LOL & $0.043 \pm 0.001$ & $0.194 \pm 0.003$ & $0.763 \pm 0.003$ \\
Super-resolution, WorldView-2 & $0.039 \pm 0.001$ & $0.237 \pm 0.003$ & $0.724 \pm 0.003$ \\
Super-resolution, OASIS3 & $0.041 \pm 0.000$ & $0.065 \pm 0.001$ & $0.894 \pm 0.001$ \\
Denoising, IXI & $0.049 \pm 0.001$ & $0.076 \pm 0.001$ & $0.875 \pm 0.002$ \\
\bottomrule
\end{tabular}
\caption{\textbf{Final loss weights set by gradient-norm matching} (share $\lambda_j/\sum_k\lambda_k$ at the end of training, mean $\pm$ s.d.\ over eight seeds backbones as in Tables~\ref{tab:apply2d} and \ref{tab:apply3d}).}
\label{tab:final_lambda}
\end{table}

Table~\ref{tab:final_lambda} lists the weights where gradient-norm matching finally settles. They are stable across seeds and follow the same order on all five tasks, in 2D and 3D: the Fourier term takes the largest share, $0.72$ to $0.89$, and the convolution the smallest, $0.04$ to $0.05$. With the gradient-norm matching, a large weight corresponds to a small raw gradient. The low weight is consistently assigned to convolution family, indicating that its raw gradient is larger than those of the other two families. Without adaptive reweighting, the convolution may dominate the optimization, providing a plausible explanation for the weaker performance of fixed uniform weighting in Table~\ref{tab:lambda}.

The weights appear to be determined by the dataset rather than the task. Denoising on SIDD removes high-frequency noise while enhancement on LOL brightens and recolors low-light images, two opposite changes to the pixels, but the two RGB tasks receive almost the same weights. The two single-channel 3D MRI datasets, super-resolution on OASIS3 and denoising on IXI, likewise share nearly the same weights, with a larger Fourier share and a smaller wavelet share than in 2D. WorldView-2, the only multispectral dataset, departs most from the RGB tasks, with the lowest convolution share ($0.039$) and the highest wavelet share ($0.237$).

\subsection{The MAdam Preconditioner}
\label{sec:madam}
Gradient-norm matching and MAdam act at different stages of each step, and they can help each other. Gradient-norm matching plays the role that loss-balancing rules such as GradNorm~\citep{chen2018gradnorm} and uncertainty weighting~\citep{kendall2018multi} play in multi-objective training: it sets the weights $\lambda_j$ that form the summed gradient $d=\sum_j\lambda_j\nabla_\theta\mathcal{L}_j$. The MAdam preconditioner~\citep{liu2026madam} then acts on $d$ before Adam~\citep{kingma2014adam} takes the step, and leaves both gradient-norm matching and Adam unchanged. It is necessary because Adam divides each update by the square root of a running average of past squared gradients when $\lambda_j$ changes during training, that average mixes gradients taken under old and new weights, so the current weights are partly averaged away~\citep{liu2026madam}. MAdam save it by keep a running average of the element-wise product of the gradients of every pair of loss terms, combines these averages with the current weights $\lambda_i\lambda_j$ into a diagonal metric $\hat M$, and passes $\hat M^{-1}d$ to Adam in place of $d$. Table~\ref{tab:lambda} varies the weighting strategy with MAdam fixed, and Table~\ref{tab:madam} turns MAdam off with the weighting strategy fixed. Algorithm~\ref{alg:rfl} shows where the two components act in one training step.

\begin{table}[t]
\centering\scriptsize
\setlength{\tabcolsep}{3pt}
\begin{tabular}{lccccc}
\toprule
Optimizer & PSNR $\uparrow$ & SSIM $\uparrow$ & LPIPS $\downarrow$ & DISTS $\downarrow$ & $\Delta$ PSNR \\
\midrule
\multicolumn{6}{l}{\emph{Image denoising, SIDD, DnCNN}} \\
MAdam + Adam & $\mathbf{40.28 \pm 0.13}$ & $0.9557 \pm 0.0012$ & $0.2488 \pm 0.0036$ & $0.1847 \pm 0.0034$ & $+0.09\ [-0.24,+0.43]$ \\
Adam alone & $40.19 \pm 0.37$ & $\mathbf{0.9559 \pm 0.0015}$ & $\mathbf{0.2463 \pm 0.0045}$ & $\mathbf{0.1825 \pm 0.0052}$ & \\
\midrule
\multicolumn{6}{l}{\emph{Image enhancement, LOL, DRBN}} \\
MAdam + Adam & $\mathbf{21.17 \pm 0.33}$ & $\mathbf{0.8035 \pm 0.0032}$ & $\mathbf{0.2568 \pm 0.0043}$ & $\mathbf{0.1995 \pm 0.0032}$ & $+0.26\ [+0.03,+0.49]$ \\
Adam alone & $20.91 \pm 0.34$ & $0.8016 \pm 0.0039$ & $0.2588 \pm 0.0043$ & $0.2018 \pm 0.0015$ & \\
\midrule
\multicolumn{6}{l}{\emph{Super-resolution, WorldView-2, SFINet}} \\
& PSNR $\uparrow$ & SSIM $\uparrow$ & ERGAS $\downarrow$ & SAM $\downarrow$ & \\
MAdam + Adam & $\mathbf{39.62 \pm 0.08}$ & $\mathbf{0.9631 \pm 0.0005}$ & $\mathbf{1.318 \pm 0.011}$ & $\mathbf{0.0297 \pm 0.0002}$ & $+0.87\ [+0.78,+0.97]$ \\
Adam alone & $38.74 \pm 0.13$ & $0.9574 \pm 0.0007$ & $1.452 \pm 0.019$ & $0.0328 \pm 0.0005$ & \\
\bottomrule
\end{tabular}
\caption{\textbf{The preconditioner, on and off} (mean $\pm$ s.d.\ across seed, eight seeds shared between the two rows of a task; bold the best value in the column of its block). ``MAdam + Adam'' is MAdam followed by Adam, as on every task in the paper. $\Delta$ PSNR is the mean per-seed difference with its $95\%$ confidence interval; MAdam + Adam is ahead on $4/8$, $7/8$ and $8/8$ seeds. The two rows of a task share the loss, the loss weights and the per-step schedule; only the preconditioner changes.}
\label{tab:madam}
\end{table}

We additionally compare MAdam against Adam alone on all three 2D image reconstruction tasks, with the loss and every other setting held fixed (Table~\ref{tab:madam}). MAdam is generally the stronger choice. It leads on every metric on low-light and pan-sharpening, by $0.26$ and $0.87$\,dB on PSNR, and is roughly level with Adam alone on denoising, where it has the higher mean PSNR, by $0.09$\,dB, while Adam alone has the better LPIPS and DISTS, by $0.003$ and $0.002$.

\subsection{Subbands of the Random Wavelet Detail}

A 3D wavelet level produces seven detail subbands and a 2D level produces three. We return the two first-order ones, LH and HL, so that the direction $u$ is uniform on the circle. Table~\ref{tab:subband} shows that on SIDD, adding the diagonal HH lowers PSNR by $0.41$\,dB, and adding both HH and the approximation LL lowers it by $0.20$\,dB. Therefore we applied the best setting empirically.

The wavelet family uses the db2 filter pair, four coefficients long. The transform is separable, uses circular padding and is cascaded across four levels without downsampling: the level-$s$ detail is computed on the level $(s{-}1)$ approximation, with the filters dilated by $2^{s-1}$. At each level we keep the first-order detail subbands, which are high-pass along one axis and low-pass along the others (LH and HL in 2D; three subbands in 3D), combine them with a random unit vector drawn independently for that level, and normalize the result to unit norm. Every draw covers all four levels. The effective support at the four levels is $4$, $10$, $22$ and $46$ pixels per axis, inside the $128\times128$ and $96^3$ training crops. Our implementation matches \texttt{pywt.swtn} coefficient by coefficient to within $2\times10^{-14}$.

\begin{table}[t]
\centering\scriptsize
\setlength{\tabcolsep}{4pt}
\begin{tabular}{lcccc}
\toprule
Wavelet detail subbands & PSNR $\uparrow$ & SSIM $\uparrow$ & LPIPS $\downarrow$ & DISTS $\downarrow$ \\
\midrule
LH, HL (ours)       & $\mathbf{40.28 \pm 0.13}$ & $\mathbf{0.9557 \pm 0.0012}$ & $\underline{0.2488 \pm 0.0036}$ & $\mathbf{0.1847 \pm 0.0034}$ \\
${}+{}$HH           & $39.87 \pm 0.41$ & $\underline{0.9555 \pm 0.0014}$ & $\mathbf{0.2481 \pm 0.0030}$ & $0.1854 \pm 0.0028$ \\
${}+{}$HH${}+{}$LL  & $\underline{40.08 \pm 0.29}$ & $0.9547 \pm 0.0007$ & $0.2505 \pm 0.0042$ & $\mathbf{0.1847 \pm 0.0034}$ \\
\bottomrule
\end{tabular}
\caption{\textbf{Which detail subbands the wavelet family returns} (image denoising, SIDD, DnCNN, eight seeds per configuration, mean $\pm$ s.d.\ across seeds). Each row changes only the subbands from the deployed configuration, which returns LH and HL. Bold is the best value in each column and underlined the second best.}
\label{tab:subband}
\end{table}

\subsection{Random Fourier Features}
\label{sec:fourier-bands}
We vary the number $D$ of random Fourier features per patch. Table~\ref{tab:width} shows that on SIDD a single feature is $0.08$ to $0.48$\,dB below the wider settings, and PSNR rises with the width from $D{=}4$ to its highest value at $D{=}64$, $40.28$\,dB. We deploy $D{=}64$ and pay for the width in FLOPs: the term is a $3^3$ convolution with $D$ output channels, which carries most of the loss's arithmetic at this setting (\S\ref{sec:cost}).

\begin{table}[t]
\centering\scriptsize
\setlength{\tabcolsep}{4pt}
\begin{tabular}{lcccc}
\toprule
Fourier width & PSNR $\uparrow$ & SSIM $\uparrow$ & LPIPS $\downarrow$ & DISTS $\downarrow$ \\
\midrule
$D=1$         & $39.80 \pm 0.44$ & $0.9537 \pm 0.0016$ & $0.2516 \pm 0.0072$ & $0.1858 \pm 0.0068$ \\
$D=4$         & $39.88 \pm 0.46$ & $0.9546 \pm 0.0017$ & $\mathbf{0.2486 \pm 0.0064}$ & $\underline{0.1848 \pm 0.0058}$ \\
$D=16$        & $\underline{40.06 \pm 0.39}$ & $\underline{0.9554 \pm 0.0010}$ & $0.2497 \pm 0.0056$ & $0.1855 \pm 0.0054$ \\
$D=64$ (ours) & $\mathbf{40.28 \pm 0.13}$ & $\mathbf{0.9557 \pm 0.0012}$ & $\underline{0.2488 \pm 0.0036}$ & $\mathbf{0.1847 \pm 0.0034}$ \\
\bottomrule
\end{tabular}
\caption{\textbf{How many random Fourier features a draw contains} (image denoising, SIDD, DnCNN, eight seeds per configuration, mean $\pm$ s.d.\ across seeds). Each row changes only the width $D$ from the deployed configuration, which uses $D{=}64$. Bold is the best value in each column and underlined the second best.}
\label{tab:width}
\end{table}

The bandwidth $\gamma$ controls the scale of the sampled projection vectors. Table~\ref{tab:gamma} sweeps $\gamma$ over five orders of magnitude. The deployed $\gamma{=}1$ records the highest mean of the six settings and the lowest standard deviation across seed, $\pm0.13$\,dB against $\pm0.17$ to $\pm0.83$, although only the interval against $\gamma{=}100$ excludes zero. The perceptual metrics prefer a slightly wider bandwidth, $\gamma{=}0.1$ leading LPIPS and DISTS by $0.002$ each.

\begin{table}[t]
\centering\small
\setlength{\tabcolsep}{5pt}
\begin{tabular}{lcccc}
\toprule
$\gamma$ & PSNR $\uparrow$ & SSIM $\uparrow$ & LPIPS $\downarrow$ & DISTS $\downarrow$ \\
\midrule
$0.01$ & $39.90 \pm 0.52$ & $0.9550 \pm 0.0017$ & $0.2491 \pm 0.0063$ & $0.1857 \pm 0.0056$ \\
$0.1$  & $\underline{40.22 \pm 0.17}$ & $0.9550 \pm 0.0014$ & $\mathbf{0.2469 \pm 0.0041}$ & $\mathbf{0.1827 \pm 0.0037}$ \\
\textbf{1 (ours)} & $\mathbf{40.28 \pm 0.13}$ & $\mathbf{0.9557 \pm 0.0012}$ & $\underline{0.2488 \pm 0.0036}$ & $\underline{0.1847 \pm 0.0034}$ \\
$10$   & $40.12 \pm 0.33$ & $\underline{0.9554 \pm 0.0008}$ & $0.2505 \pm 0.0061$ & $0.1860 \pm 0.0053$ \\
$100$  & $39.46 \pm 0.80$ & $0.9549 \pm 0.0008$ & $0.2562 \pm 0.0043$ & $0.1923 \pm 0.0040$ \\
$1000$ & $39.74 \pm 0.83$ & $0.9543 \pm 0.0008$ & $0.2518 \pm 0.0054$ & $0.1891 \pm 0.0069$ \\
\bottomrule
\end{tabular}
\caption{\textbf{The bandwidth of the random Fourier family} (SIDD, DnCNN, eight seeds per configuration, mean $\pm$ s.d.\ across seeds). The three families and everything else are held fixed and only $\gamma$ changes, so the $\gamma{=}1$ row is the deployed configuration of Table~\ref{tab:threeq}. Bold is the best value in each column and underlined the second best.}
\label{tab:gamma}
\end{table}

\paragraph{Frequency-band analysis.}
We partition the SIDD test error into low ($|f|<1/8$), mid ($1/8\le |f|<1/4$), and high ($|f|\ge 1/4$) radial DFT bands, with frequencies in cycles per pixel. We compute PSNR from the error restricted to each band for each test image and average over images. The comparison uses the same eight seeds for \method{} and its Convolution + Wavelet ablation.

\begin{center}
\scriptsize
\setlength{\tabcolsep}{4pt}
\begin{tabular}{lrrrrrrrr}
\toprule
& \multicolumn{4}{c}{With Fourier (dB)} & \multicolumn{4}{c}{Without Fourier (dB)} \\
\cmidrule(lr){2-5}\cmidrule(lr){6-9}
& Low & Mid & High & Full & Low & Mid & High & Full \\
\midrule
Mean & 42.301 & 46.937 & 49.145 & 40.281 & 42.108 & 46.478 & 48.307 & 39.880 \\
\bottomrule
\end{tabular}
\end{center}

Removing the Fourier family lowers mean PSNR across all three bands ($-0.19$\,dB for low, $-0.46$\,dB for mid, and $-0.84$\,dB for high), with the largest degradation in the high-frequency band. Fig.~\ref{fig:qualfourierapp} shows a visual example.
\subsection{Randomization schedule on all image reconstruction tasks}

\begin{table}[t]
\centering\scriptsize
\setlength{\tabcolsep}{3pt}
\begin{tabular}{llcccc}
\toprule
Task & Schedule & PSNR $\uparrow$ & SSIM $\uparrow$ & LPIPS $\downarrow$ & DISTS $\downarrow$ \\
\midrule
\multirow{3}{*}{SIDD, DnCNN} & every step & $\mathbf{40.28 \pm 0.13}$ & $\mathbf{0.9557 \pm 0.0012}$ & $\mathbf{0.2488 \pm 0.0036}$ & $\mathbf{0.1847 \pm 0.0034}$ \\
& every epoch & $\underline{39.78 \pm 0.67}$ & $0.9545 \pm 0.0014$ & $0.2503 \pm 0.0054$ & $0.1863 \pm 0.0068$ \\
& one draw    & $39.67 \pm 1.03$ & $\underline{0.9546 \pm 0.0011}$ & $\underline{0.2494 \pm 0.0052}$ & $\underline{0.1852 \pm 0.0056}$ \\
\midrule
\multirow{3}{*}{LOL, DRBN} & every step & $\mathbf{21.17 \pm 0.33}$ & $\underline{0.8035 \pm 0.0032}$ & $0.2568 \pm 0.0043$ & $0.1995 \pm 0.0032$ \\
& every epoch & $\underline{20.86 \pm 0.38}$ & $\mathbf{0.8197 \pm 0.0052}$ & $\mathbf{0.2352 \pm 0.0045}$ & $\mathbf{0.1841 \pm 0.0037}$ \\
& one draw    & $19.55 \pm 4.72$ & $0.7304 \pm 0.2330$ & $\underline{0.2994 \pm 0.1406}$ & $\underline{0.2201 \pm 0.0741}$ \\
\midrule
& & PSNR $\uparrow$ & SSIM $\uparrow$ & ERGAS $\downarrow$ & SAM $\downarrow$ \\
\multirow{3}{*}{WV2, SFINet} & every step & $\mathbf{39.62 \pm 0.08}$ & $\mathbf{0.9631 \pm 0.0005}$ & $\mathbf{1.318 \pm 0.011}$ & $\mathbf{0.0297 \pm 0.0002}$ \\
& every epoch & $39.54 \pm 0.08$ & $\underline{0.9628 \pm 0.0006}$ & $1.321 \pm 0.014$ & $\underline{0.0299 \pm 0.0003}$ \\
& one draw    & $\underline{39.55 \pm 0.12}$ & $0.9627 \pm 0.0009$ & $\underline{1.319 \pm 0.019}$ & $\underline{0.0299 \pm 0.0004}$ \\
\midrule
& & PSNR $\uparrow$ & SSIM $\uparrow$ & LPIPS $\downarrow$ & DISTS $\downarrow$ \\
\multirow{3}{*}{OASIS3, SwinUNETR} & every step & $\mathbf{22.22 \pm 0.03}$ & $\mathbf{0.8249 \pm 0.0017}$ & $\underline{0.2962 \pm 0.0010}$ & $\underline{0.2547 \pm 0.0010}$ \\
& every epoch & $22.16 \pm 0.03$ & $0.8187 \pm 0.0110$ & $0.2972 \pm 0.0014$ & $0.2554 \pm 0.0012$ \\
& one draw    & $\underline{22.21 \pm 0.05}$ & $\underline{0.8246 \pm 0.0024}$ & $\mathbf{0.2954 \pm 0.0011}$ & $\mathbf{0.2545 \pm 0.0007}$ \\
\midrule
\multirow{3}{*}{IXI-MRA, 3D DnCNN} & every step & $\mathbf{29.66 \pm 0.19}$ & $\mathbf{0.9627 \pm 0.0007}$ & $\mathbf{0.0646 \pm 0.0012}$ & $\mathbf{0.0785 \pm 0.0013}$ \\
& every epoch & $\underline{29.09 \pm 1.12}$ & $\underline{0.9616 \pm 0.0013}$ & $\underline{0.0664 \pm 0.0033}$ & $\underline{0.0811 \pm 0.0058}$ \\
& one draw    & $26.42 \pm 9.43$ & $0.8066 \pm 0.4429$ & $0.1195 \pm 0.1570$ & $0.1287 \pm 0.1446$ \\
\bottomrule
\end{tabular}
\caption{\textbf{The randomization schedule on all five tasks, every metric} (mean $\pm$ s.d.\ across seeds). The three rows of a task share the loss, the loss weights and the optimizer and differ only in the frequency of draw. The SIDD block is the Q2 block of Table~\ref{tab:threeq}.}
\label{tab:sched_full}\label{tab:sched}
\end{table}

By the column means a per-step draw is the highest on PSNR on all five tasks.
By the paired difference it is the best frequency for draw, by $0.06$ to $0.57$\,dB. On WorldView 2 both margins are about $0.07$\,dB. Its standard deviation across seed is the smallest on three of the five. On OASIS3 and WorldView-2 the per-epoch draw is tighter, by $0.009$ and $0.003$\,dB.
On LOL and IXI one held seed does not train at all, at $7.9$ and $3.1$\,dB. On LOL the per-epoch draw leads SSIM, LPIPS and DISTS, by $0.016$, $0.022$ and $0.015$; on the other four tasks the per-step draw leads every metric or is within the standard deviation of the leader.

\subsection{Transfer of the SIDD-selected default}
\label{sec:transfer-ablations}

We repeat the filter-family and initialization ablations on WorldView-2 with SFINet using the settings of Table~\ref{tab:apply2d}. Fourier + Wavelet reaches similar mean PSNR to the default, whose cross-seed PSNR spread is the smallest (Table~\ref{tab:wv2-ablation}). The four initializations differ by at most $0.05$\,dB in mean PSNR on WorldView-2.

\begin{table}[t]
\centering\scriptsize
\setlength{\tabcolsep}{3pt}
\resizebox{\linewidth}{!}{%
\begin{tabular}{lccccc}
\toprule
Configuration & \shortstack{SIDD\\PSNR $\uparrow$} & \shortstack{WV2\\PSNR $\uparrow$} & SSIM $\uparrow$ & ERGAS $\downarrow$ & SAM $\downarrow$ \\
\midrule
Default (all three, Gaussian) & $40.28 \pm 0.13$ & $39.62 \pm 0.08$ & $0.9631 \pm 0.0005$ & $1.318 \pm 0.011$ & $0.0297 \pm 0.0002$ \\
\midrule
\multicolumn{6}{l}{\emph{Filter families (Gaussian initialization)}} \\
Fourier + Wavelet & $39.28 \pm 1.36$ & $39.64 \pm 0.15$ & $0.9634 \pm 0.0012$ & $1.304 \pm 0.024$ & $0.0295 \pm 0.0005$ \\
Fourier only & $39.92 \pm 0.32$ & $39.47 \pm 0.13$ & $0.9629 \pm 0.0005$ & $1.338 \pm 0.014$ & $0.0300 \pm 0.0004$ \\
Conv. + Fourier & $39.98 \pm 0.33$ & $39.41 \pm 0.13$ & $0.9626 \pm 0.0008$ & $1.350 \pm 0.019$ & $0.0303 \pm 0.0005$ \\
Conv. + Wavelet & $39.88 \pm 0.75$ & $39.28 \pm 0.13$ & $0.9608 \pm 0.0008$ & $1.365 \pm 0.018$ & $0.0309 \pm 0.0004$ \\
Conv. only & $39.75 \pm 0.17$ & $38.20 \pm 0.20$ & $0.9544 \pm 0.0019$ & $1.582 \pm 0.054$ & $0.0351 \pm 0.0009$ \\
Wavelet only & $32.26 \pm 3.13$ & $34.74 \pm 4.15$ & $0.9563 \pm 0.0064$ & $2.229 \pm 1.372$ & $0.0685 \pm 0.0474$ \\
\midrule
\multicolumn{6}{l}{\emph{Initialization (all three families)}} \\
Kaiming normal & $39.92 \pm 0.60$ & $39.57 \pm 0.12$ & $0.9631 \pm 0.0009$ & $1.321 \pm 0.019$ & $0.0298 \pm 0.0004$ \\
Xavier uniform & $39.98 \pm 0.41$ & $39.60 \pm 0.18$ & $0.9631 \pm 0.0012$ & $1.318 \pm 0.026$ & $0.0297 \pm 0.0006$ \\
Sparse & $39.91 \pm 0.31$ & $39.59 \pm 0.12$ & $0.9632 \pm 0.0008$ & $1.318 \pm 0.017$ & $0.0297 \pm 0.0004$ \\
\bottomrule
\end{tabular}}
\caption{\textbf{Filter-family and initialization ablations on WorldView-2 with SFINet.} The SIDD PSNR column repeats Table~\ref{tab:threeq}. Remaining metrics are for WorldView-2. Values are mean $\pm$ s.d.\ over eight seeds.}
\label{tab:wv2-ablation}
\end{table}

On SIDD, we also evaluate the unchanged default against pixel $\ell_1$ on U-Net (learning rate $10^{-3}$) and blind DRUNet without a noise-level map (learning rate $10^{-4}$); \method{} matches or slightly improves mean PSNR ($+0.24$\,dB on U-Net and $+0.04$\,dB on DRUNet) while consistently lowering LPIPS and DISTS for both backbones (Table~\ref{tab:sidd-backbone-transfer}).

\begin{table}[t]
\centering\scriptsize
\setlength{\tabcolsep}{4pt}
\begin{tabular}{llcccc}
\toprule
Backbone & Loss & PSNR $\uparrow$ & SSIM $\uparrow$ & LPIPS $\downarrow$ & DISTS $\downarrow$ \\
\midrule
DnCNN (Table~\ref{tab:apply2d}) & pixel $\ell_1$ & $39.64 \pm 1.00$ & $0.9551 \pm 0.0011$ & $0.2478 \pm 0.0056$ & $0.1830 \pm 0.0057$ \\
 & \method{} & $40.28 \pm 0.13$ & $0.9557 \pm 0.0012$ & $0.2488 \pm 0.0036$ & $0.1847 \pm 0.0034$ \\
\midrule
U-Net & pixel $\ell_1$ & $35.21 \pm 0.65$ & $0.9079 \pm 0.0022$ & $0.3406 \pm 0.0075$ & $0.2414 \pm 0.0031$ \\
 & \method{} & $35.45 \pm 0.89$ & $0.9208 \pm 0.0074$ & $0.3090 \pm 0.0071$ & $0.2253 \pm 0.0025$ \\
\midrule
DRUNet & pixel $\ell_1$ & $42.02 \pm 0.03$ & $0.9669 \pm 0.0001$ & $0.2215 \pm 0.0009$ & $0.1873 \pm 0.0007$ \\
 & \method{} & $42.06 \pm 0.05$ & $0.9669 \pm 0.0002$ & $0.2197 \pm 0.0007$ & $0.1863 \pm 0.0007$ \\
\bottomrule
\end{tabular}
\caption{\textbf{The SIDD-selected default across backbones on SIDD.} Values are mean $\pm$ s.d.\ over eight seeds ($42$--$49$).}
\label{tab:sidd-backbone-transfer}
\end{table}
\section{Qualitative Results}
\label{sec:qual2d}
The figures in the main paper show one example per task. This section provides additional examples, five to nine per task, using the same configurations and held-out splits (Figs.~\ref{fig:qualsiddapp} to \ref{fig:qualixiapp}).
\begin{figure}[htbp]
\centering
\includegraphics[width=\linewidth]{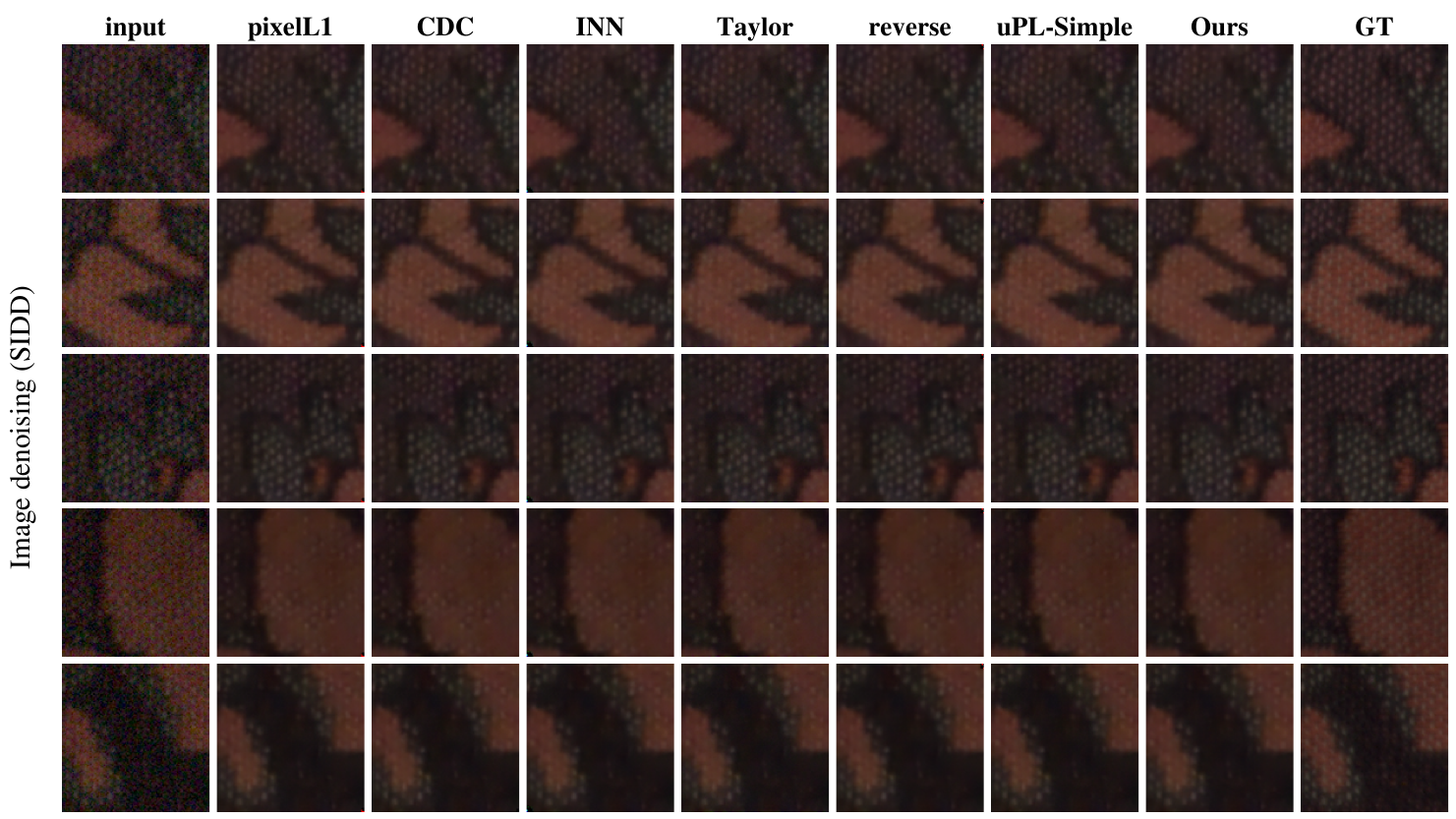}
\caption{\textbf{Image denoising on SIDD, five held-out patches.} The untrained perceptual loss is shown in its SimpleNet form (uPL-Simple).}
\label{fig:qualsiddapp}
\end{figure}

\begin{figure}[htbp]
\centering
\includegraphics[width=\linewidth]{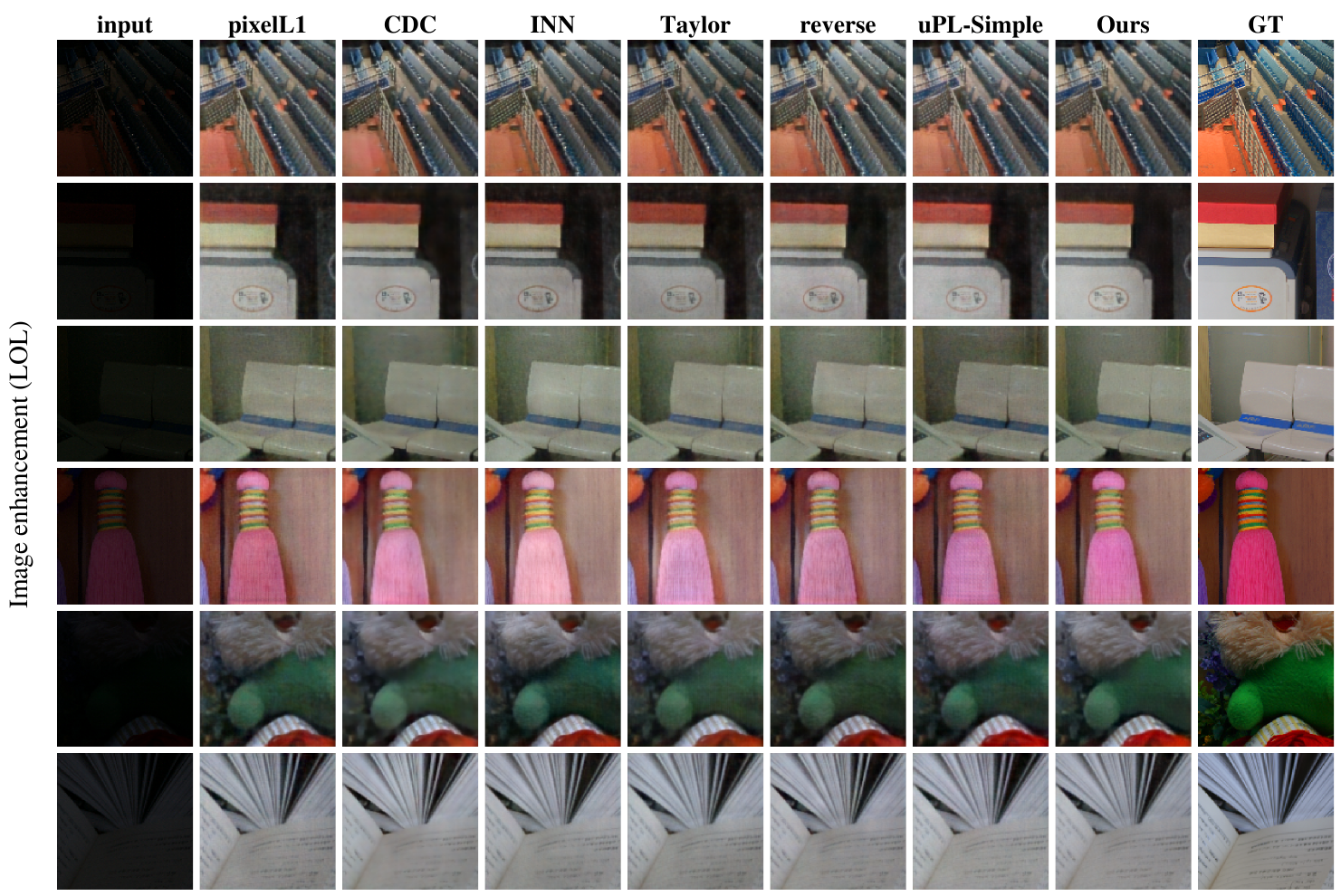}
\caption{\textbf{Image enhancement on LOL, six held-out images.} This is the task where the losses differ in a way the eye can judge, exposure and color cast rather than sub-decibel texture.}
\label{fig:quallolapp}
\end{figure}

\begin{figure}[htbp]
\centering
\includegraphics[width=\linewidth]{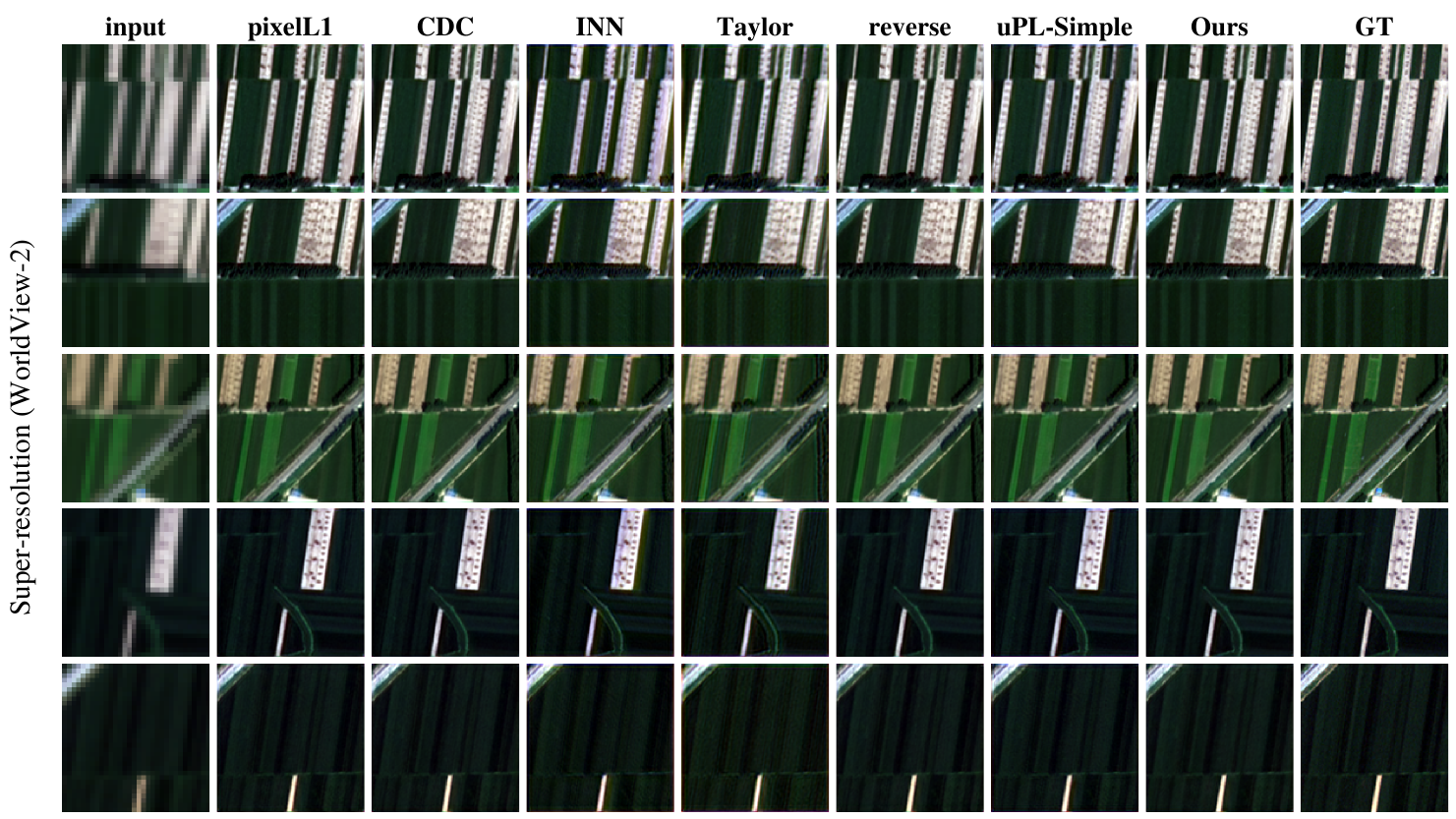}
\caption{\textbf{Super-resolution on WorldView-2, five held-out patches.} The backbone is SFINet, as in Table~\ref{tab:apply2d}.}
\label{fig:qualwv2app}
\end{figure}

\begin{figure}[htbp]
\centering
\includegraphics[width=\linewidth]{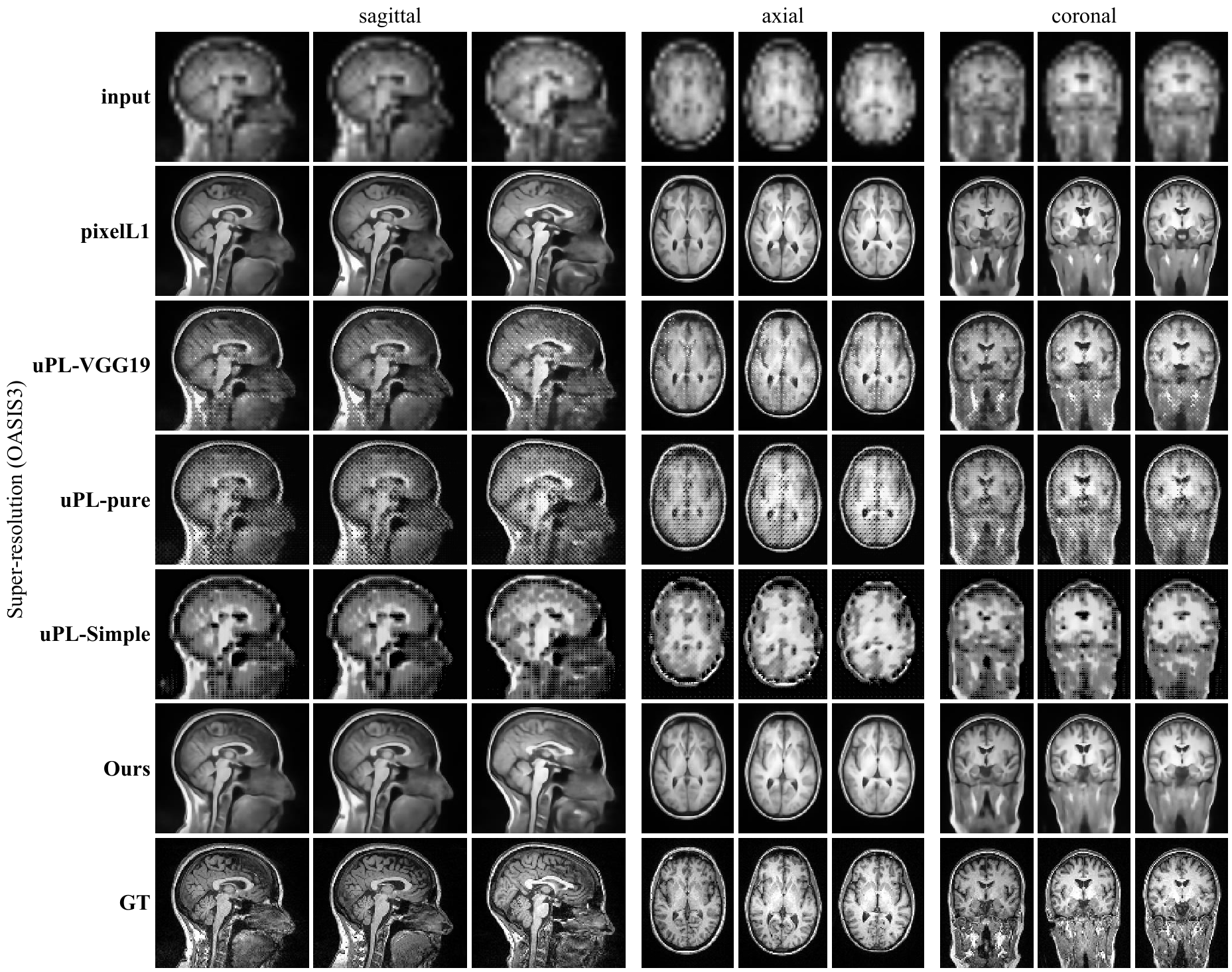}
\caption{\textbf{Super-resolution on OASIS3, nine held-out slices, three sagittal, three axial and three coronal.} Rows are the losses of Fig.~\ref{fig:qual3d}, in the same order. The lattice the three untrained-network variants lay over the parenchyma is present on every slice and in every instantiation.}
\label{fig:qualoasis3app}
\end{figure}

\begin{figure}[htbp]
\centering
\includegraphics[width=\linewidth]{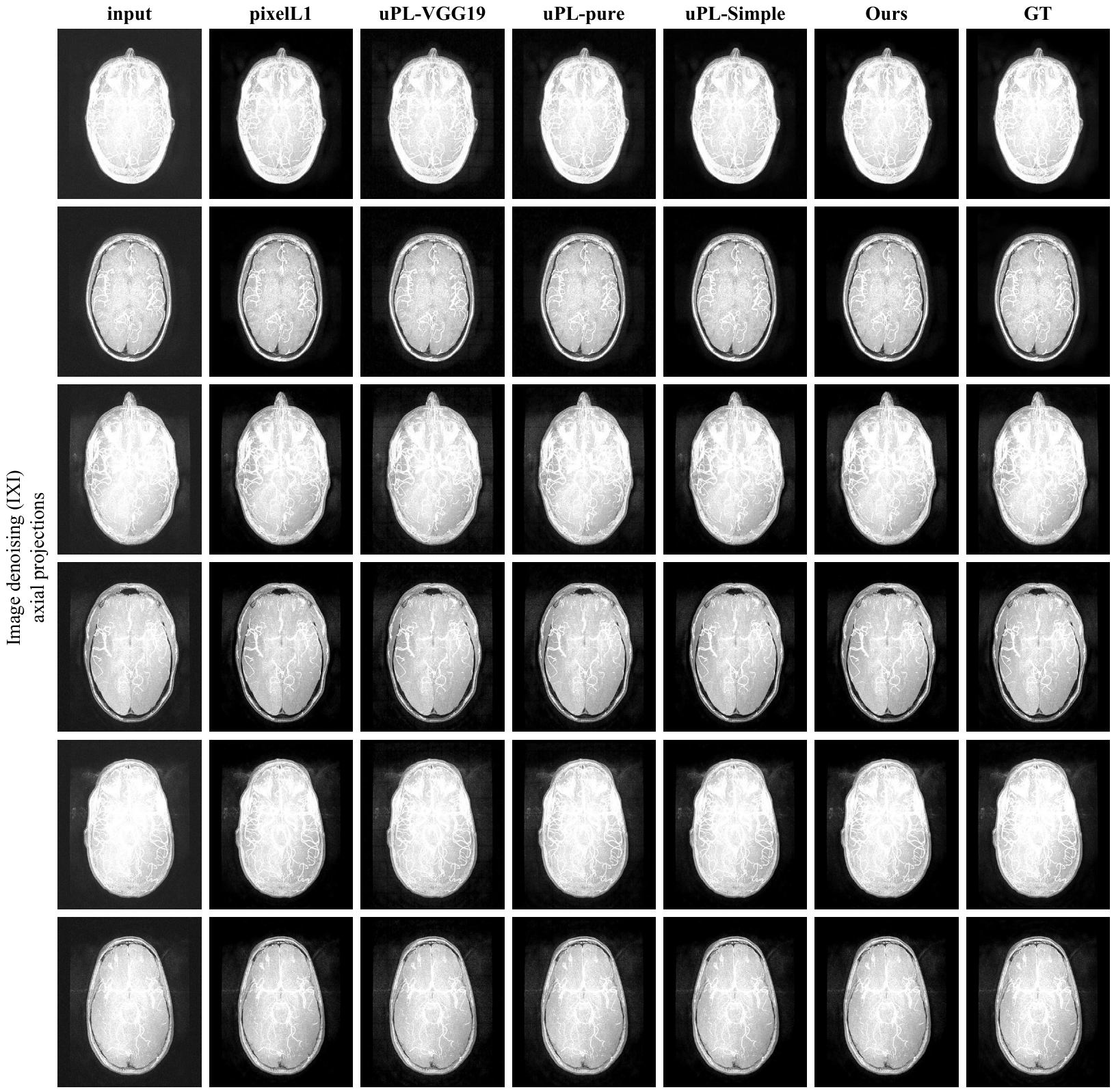}
\caption{\textbf{Image denoising on IXI, six held-out volumes, axial slab projections.} Columns as in Fig.~\ref{fig:qual3d}. This is the 3D task on which \method{} and the pixel loss are level in Table~\ref{tab:apply3d} and \method{} keep the vessel continuity better than the untrained network loses.}
\label{fig:qualixiapp}
\end{figure}

\begin{figure}[htbp]
\centering
\includegraphics[width=0.7\linewidth]{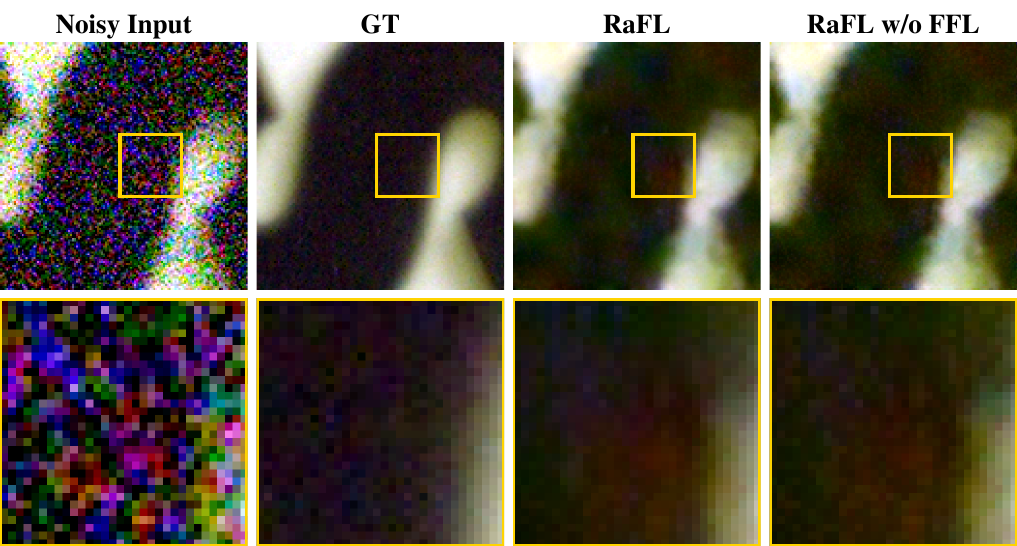}
\caption{\textbf{Image denoising on SIDD with and without the Fourier family, one example patch.} From left to right: noisy input, ground truth, \method{}, and \method{} without the random Fourier features (Convolution + Wavelet; ``RaFL w/o FFL'' in the panel title). The bottom row enlarges the boxed region. Without the Fourier family, a short stretch of the edge in the enlarged region is greener than in the \method{} output and in the ground truth.}
\label{fig:qualfourierapp}
\end{figure}

\section{Implementation details}
\label{sec:implementation}\label{sec:instantiation}

\subsection{Preprocessing}
\label{sec:preproc}

All intensities are scaled to $[0,1]$. 2D images are divided by their maximum value $255$ for the 8-bit SIDD and LOL images, and $2047$ for the 11-bit WorldView-2 images. The 3D volume is clipped to its 1st and 99th intensity percentiles (OASIS3) or to its 99.5th percentile (IXI), which removes a few extreme voxels, and then rescaled to $[0,1]$.
Training uses $128\times128$ patches on SIDD, LOL and WorldView-2, whole $80\times128\times128$ volumes on OASIS3 and random $96^3$ crops on IXI; the only augmentation is a random horizontal flip on SIDD dataset. The WorldView-2 uses the reduced-resolution pairs provided with the dataset. The OASIS3 input is made by $4\times$ trilinear downsampling and upsampling, and the IXI input by adding Rician noise with $\sigma=0.05$, drawn anew in training and fixed per volume at test time.
For 2D comparison table (Table~\ref{tab:apply2d}) SIDD: DnCNN, $60$ epochs, eight seeds per configuration; LOL: DRBN, $30$ epochs, eight seeds per configuration; WorldView-2: SFINet, $30$ epochs, eight seeds per configuration, $145$ held-out patches mean $\pm$ s.d.\ across seeds, bold the best value in the column of its block). uPL is the SimpleNet recommended by \citet{pfaehler2025untrained}, with no pixel term. CDC, reverse, Taylor and INN are the four prototypes of \citet{zhou2023training}, each added to the pixel loss as in that paper.
For 3D comparison table (Table~\ref{tab:apply3d}) OASIS3: T1w, SwinUNETR, $12$ epochs, $137$ held-out scans; IXI: $5\%$ Rician noise, 3D DnCNN, $12$ epochs; eight seeds per configuration, mean $\pm$ s.d.\ across seeds, bold the best value in the column of its block). uPL is the 3D SimpleNet recommended by \citet{pfaehler2025untrained}, with no pixel term.

\subsection{Algorithm}
Algorithm~\ref{alg:rfl} shows one training step. The three filter families are redrawn, applied with the same draw to the prediction and the target, then compared with an $\ell_1$ distance (without a pixel term). Every $50$ steps, gradient-norm matching resets the loss weights $\lambda_j$ (App.~\ref{sec:balance}), and MAdam preconditions the weighted gradient before the Adam update (App.~\ref{sec:madam}). The random weight are discarded after the step, so nothing is added to the network at inference.

\begin{algorithm}[h!]
\caption{One \method{} training step}
\label{alg:rfl}
\begin{algorithmic}[1]
\STATE Sample minibatch $(x,y)$; predict $\hat{y}=f_\theta(x)$
\STATE \textbf{Redraw} the three random filters $\phi_j$, $j=1,2,3$ \hfill\COMMENT{a new draw at every step}
\STATE Compute the three terms $\mathcal{L}_j=\|\phi_j(\hat{y})-\phi_j(y)\|_1$
\STATE Every $50$ steps, reset $\lambda_j$ so that $\|\lambda_j\nabla_{\hat y}\mathcal{L}_j\|$ is equal across $j$ \hfill\COMMENT{gradient-norm matching, on the output}
\STATE Compute $g=\nabla_\theta\sum_j\lambda_j\mathcal{L}_j$ \hfill\COMMENT{one backward pass} \STATE Refresh the MAdam statistics of one randomly drawn pair of terms from their parameter gradients \hfill\COMMENT{at most two extra backward passes}
\STATE $\theta\leftarrow\text{optimizer-step}(\theta, g)$ \hfill\COMMENT{preconditioned by MAdam, then Adam, on every task; see \S\ref{sec:optim}}
\end{algorithmic}\label{algorithm}
\end{algorithm}

\subsection{Loss Complexity}\label{apdx:lossComplexity}

\begin{table}[t] \centering\small \setlength{\tabcolsep}{5pt} \begin{tabular}{lrrr@{\hspace{14pt}}lrrr}
\toprule
\multicolumn{4}{c}{\emph{Image denoising, SIDD, DnCNN}} & \multicolumn{4}{c}{\emph{Super-resolution, OASIS3, SwinUNETR}} \\
\cmidrule(lr){1-4}\cmidrule(lr){5-8}
Loss & GFLOPs & of fwd & ms/it & Loss & GFLOPs & of fwd & ms/it \\
\midrule
\textbf{\method{}} & $1.12$& $1.9\%$ & $8.3$& \textbf{\method{}} & $11.8$& $1.0\%$ & $265.2$\\
\method{}, Adam only & $1.12$& $1.9\%$ & $4.6$& \method{}, Adam only & $11.8$ & $1.0\%$ & $135.6$\\
pixel $\ell_1$ & $0.00$ & $0.0\%$ & 	$3.3$& pixel $\ell_1$ & $0.00$ & $0.0\%$ & $128.1$\\
CDC & $5.00$ & $7.8\%$ & $4.2$& CDC & $42.3$ & $3.5\%$ & $132.8$\\
reverse & $0.31$ & $0.5\%$ & $3.6$& reverse & $28.8$ & $2.4\%$ & $209.5$\\
Taylor & $0.21$ & $0.4\%$ & $3.6$& Taylor & $0.71$ & $0.1\%$ & $131.5$\\
INN & $0.05$ & $0.1\%$ & $3.7$& INN & $0.85$ & $0.1\%$ & $131.4$\\
uPL, SimpleNet & $10.12$ & $14.7\%$ & $3.6$& uPL, SimpleNet & $294.4$ & $20.0\%$ & $133.8$\\
uPL, VGG19 & $203.8$ & $77.6\%$ & $5.4$& uPL, VGG19 & $1821.0$ & $60.7\%$ & $138.4$\\
\bottomrule
\end{tabular}
\caption{Computational cost of the loss functions on SIDD and OASIS3. GFLOPs measures the loss term only. of fwd denotes the fraction of combined loss and backbone forward FLOPs, and ms/it is the time of one training step, measured with Adam for every baseline.}
\label{tab:cost}\label{tab:cost2d3d} \end{table}

FLOPs in Table~\ref{tab:cost} are counted with the \texttt{FlopCounterMode} of PyTorch on a single forward pass of the loss terms, prediction and target both, at the batch each task trains with; the counter covers convolution, matrix multiplication and attention and not elementwise work, on every loss alike.

Profiling our original SIDD implementation found about $830$ CUDA kernel launches per step, so we refactored the loss. The refactored implementation applies the convolution and wavelet filters, which are linear and bias-free, to the residual, batches the wavelet convolutions in the frequency domain, and evaluates the nonlinear Fourier features for prediction and target in one batch. It also caches the MAdam pair weights between loss-weight updates and reuses the loss-term output gradients across auxiliary passes. In double precision, the refactored loss matches the original one to within $10^{-14}$. With it, a training step on SIDD takes $8.3$\,ms with MAdam and $4.6$\,ms with Adam alone, against $4.2$\,ms for CDC in Table~\ref{tab:cost}.

GFLOPs is the loss term alone on one forward pass, prediction and target both; ``of fwd'' is its share of the loss plus the backbone's own forward pass, $58.9$ GFLOPs for DnCNN and $1176.5$ for SwinUNETR, a ratio of forward FLOPs and not of a whole training step; ``ms/it'' is the training step in milliseconds.
The \method{} Adam only row removes the MAdam preconditioner. The GFLOPs columns count the loss alone and leave out the preconditioner, whose cost appears only in the step time. Inference is $0.70$\,ms on every SIDD row and $41.9$\,ms on every OASIS3 row, because the deployed network is the backbone, unchanged.

\fi

\end{document}